\documentclass[pdflatex,sn-basic]{sn-jnl}

\usepackage{graphicx}%
\usepackage{multirow}%
\usepackage{amsmath,amssymb,amsfonts}%
\usepackage{amsthm}%
\usepackage{mathrsfs}%
\usepackage[title]{appendix}%
\usepackage{xcolor}%
\usepackage{textcomp}%
\usepackage{manyfoot}%
\usepackage{booktabs}%
\usepackage[ruled,vlined,linesnumbered]{algorithm2e}
\usepackage{listings}%
\usepackage{epsfig}
\usepackage{subcaption}
\usepackage{calc}
\usepackage{multicol}
\usepackage{pslatex}
\usepackage{verbatim}
\usepackage{cuted}
\usepackage[bottom]{footmisc}
\usepackage[ruled,vlined,linesnumbered]{algorithm2e}

\usepackage{hyperref}
\usepackage{cleveref}

\definecolor{codegreen}{rgb}{0,0.6,0}
\definecolor{codegray}{rgb}{0.5,0.5,0.5}
\definecolor{codepurple}{rgb}{0.58,0,0.82}
\definecolor{backcolour}{rgb}{0.97,0.97,0.97}

\lstdefinestyle{python_jay}{
    backgroundcolor=\color{backcolour},
    commentstyle=\color{codegreen},
    keywordstyle=\color{blue},
    numberstyle=\tiny\color{codegray},
    stringstyle=\color{codepurple},
    basicstyle=\ttfamily\footnotesize,
    breakatwhitespace=false,
    breaklines=true,
    captionpos=b,
    keepspaces=true,
    numbers=left,
    numbersep=5pt,
    showspaces=false,
    showstringspaces=false,
    showtabs=false,
    tabsize=2,
}

\begin{document}

\title[Article Title]{Rock the KASBA: Blazingly Fast and Accurate Time Series Clustering}


\author[1]{\fnm{Christopher} \sur{Holder}}\email{c.l.holder@soton.ac.uk}

\author[1,2]{\fnm{Anthony} \sur{Bagnall}}\email{a.j.bagnall@soton.ac.uk}
\equalcont{These authors contributed equally to this work.}

\affil[1]{\orgdiv{School of Electronics and Computer Science}, \orgname{University of Southampton}, \orgaddress{\street{University Rd}, \city{Southampton}, \postcode{SO17 1BJ}, \state{Hampshire}, \country{England}}}
\affil[2]{\orgdiv{School of Computing Sciences, University}, \orgname{University of East Anglia}, \orgaddress{\city{Norwich}, \postcode{NR4 7TJ}, \state{Norfolk}, \country{England}}}

\abstract{Time series data has become increasingly prevalent across numerous domains, driving a growing demand for time series machine learning techniques. Among these, time series clustering (TSCL) stands out as one of the most popular machine learning tasks. TSCL serves as a powerful exploratory analysis tool and is also employed as a preprocessing step or subroutine for various tasks, including anomaly detection, segmentation, and classification. 

The most popular TSCL algorithms are either fast (in terms of run time) but perform poorly on benchmark problems, or perform well on benchmarks but scale poorly. We present a new TSCL algorithm, the $k$-means (K) accelerated (A) Stochastic subgradient (S) Barycentre (B) Average (A) (KASBA) clustering algorithm. KASBA is a $k$-means clustering algorithm that uses the Move-Split-Merge (MSM) elastic distance at all stages of clustering, applies a randomised stochastic subgradient gradient descent to find barycentre centroids, links each stage of clustering to accelerate convergence and exploits the metric property of MSM distance to avoid a large proportion of distance calculations. It is a versatile and scalable clusterer designed for real-world TSCL applications. It allows practitioners to balance  run time and clustering performance. We demonstrate through extensive experimentation that KASBA produces significantly better clustering than the faster state of the art clusterers and is offers orders of magnitude improvement in run time over the most performant $k$-means alternatives.}




\maketitle

\section{Introduction}
We consider a time series as an ordered sequence of real valued observations. Time series data has become ubiquitous, emerging across numerous domains such as astronomy, biology, engineering, finance, manufacturing, medicine, meteorology, and more~\cite{mcdowell18tsclbiology,wenig24JET,villeta23metorologyuse}. The widespread generation of time series data, coupled with the desire to analyse and derive insights from it, has driven substantial interest in time series machine learning tasks such as anomaly detection, classification, clustering, forecasting, querying, regression, and segmentation. One of the most popular fields is time series clustering (TSCL)~\cite{holder24review,lafabregue22deep}. The objective of TSCL is to group time series into clusters where the series within a cluster exhibit homogeneity, while those in different clusters display heterogeneity. TSCL is a common starting point for exploratory data analysis~\cite{zolhavarieh14subsequent} and is also commonly used as a step in a supervised pipeline~\cite{dhariyal23channel}.

A collection of equal length univariate time series sampled at equal frequency can be treated as tabular data, where each column is a time point. This means a standard clustering algorithm can be applied. However, ignoring the ordering inherent in the data will usually result in a worse model. There have been many time series specific clustering algorithms proposed. These can broadly be grouped into deep learning~\cite{alqahtani21deepclusteringreview}, feature based~\cite{zhang19ussl} and distance based algorithms~\cite{holder24review}. Historically, distance based algorithms have been the most popular and are the focus of this work. There have been a large number of time series specific distance functions proposed and numerous comparative studies of their application to TSCL~\cite{holder24review} and classification~\cite{lines15elastic}. These have empirically shown that distance measures that ignore temporal ordering (e.g., Euclidean distance) yield significantly worse similarity measures between time series for these tasks. This is because small misalignments can lead to large distance scores for series that are conceptually similar in shape. The most widely used distance measure that compensates for offset is Dynamic Time Warping (DTW)~\cite{ratanamahatana05threemyths}, the first in a family of algorithms commonly known as elastic distances~\cite{lines15elastic}. Elastic distances account for misalignment between time series during distance computation.

The most common approach to clustering time series data is to use traditional clustering algorithms but to replace the standard distance measure with a time series distance measure. Among these, DTW has been the most widely used for adapting traditional clustering algorithms such as $k$-means~\cite{petitjean11dba}, $k$-medoids~\cite{holder23kmedoids}, Agglomerative Clustering~\cite{fadi22agglodtw}, Density Peaks~\cite{begum16tadpoles}, and DBSCAN~\cite{javed20benchmark}. For many years, DTW combined with traditional clustering models, such as Partitioning Around Medoids (PAM) and Agglomerative Clustering, was considered state-of-the-art~\cite{javed20benchmark}. More recent algorithms such as $k$-Shape~\cite{paparrizos16kshapes}, $k$-means using DTW barycentre averaging (DBA)~\cite{petitjean11dba} soft-DBA~\cite{cuturi2017softdtw} and shape-DBA~\cite{ali23shapedba} improve TSCL performance within a $k$-means clustering framework. However, with the exception of $k$-Shape, these algorithms require significant computation, thereby limiting their applicability for real-world TSCL applications. This underscores that clustering performance is not the sole consideration for TSCL practitioners; the run time of an algorithm is of similar importance for many applications. 

Our focus is on identifying the most effective approach to applying $k$-means for TSCL. To this end, we aim to bridge the gap between the performance of shape-DBA and the speed of $k$-Shape by introducing KASBA: the $k$-means (K) Accelerated (A) Stochastic subgradient (S) Barycentre (B) Average (A) clustering algorithm. KASBA is designed to work seamlessly with any elastic distance that satisfies the properties of a metric, integrating it into every stage of the $k$-means process to create a fully end-to-end solution. While this paper focuses on experiments using the Move-Split-Merge (MSM)~\cite{stefan13msm} elastic distance, which is a metric, the algorithm is flexible. Practitioners who wish to use another elastic distance metric, such as TWE~\cite{marteau09twe}, can easily substitute the appropriate function in the implementation.

For initialisation, we adapt k-means++~\cite{arthur07kmeansplusplus} to use an elastic distance instead of the Euclidean distance. In the assignment phase, we exploit the distance measures metric property to reduce the number of distance calculations using the triangle inequality~\cite{elkan03elkankmeans}. During the averaging stage, we employ barycentre averaging~\cite{petitjean11dba} made generic for any elastic distance~\cite{holder23mba} combined with a form of subgradient descent inspired by~\cite{schultz18ssgba}. Further algorithmic details are provided in Section~\ref{sec:kasba}.

We evaluate KASBA through an extensive experimental evaluation, comparing its clustering performance and run time against a range of current state-of-the-art clustering algorithms. Our results demonstrate that KASBA delivers state-of-the-art performance in a fraction of the time of the most popular TSCL algorithms. we perform an ablation study to evaluate the contributions of KASBA’s structural components and parameters, offering deeper insights into the mechanisms behind its efficiency.

This paper is structured as follows. Section~\ref{sec:preliminaries} provides background into TSCL and reviews related research in TSCL such as elastic distances and barycentre averaging. Section~\ref{sec:kasba} describes the KASBA algorithm in detail and identifies how KASBA combines and extends a range of TSCL $k$-means related TSCL research. We describe our experimental set up and present results in Section~\ref{sec:results} before a closer exploration of KASBA performance in Section~\ref{sec:analysis} and concluding in Section~\ref{sec:conclusions}. KASBA is available as scikit-learn compatible estimator in the \texttt{aeon} toolkit~\cite{aeon24jmlr}\footnote{\url{https://aeon-toolkit.org/}} as are most of the other clustering algorithms we use. All our experiments are conducted with the \texttt{aeon} toolkit and are easily reproducible using the python package \texttt{tsml-eval}\footnote{\url{https://github.com/time-series-machine-learning/tsml-eval}}. All results, further analysis and details on reproducing experiments are provided in the accompanying notebook\footnote{\url{https://github.com/time-series-machine-learning/tsml-eval/tree/main/tsml_eval/publications/clustering/kasba/kasba.ipynb}}.

\section{Background}
\label{sec:preliminaries}

A time series is sequence of $m$ ordered real valued observations of a variable, 
$\mathbf{x}=(x_1,\dots,x_m)$.
If each observation $x_i$ is a vector, $x$ is a multivariate time series. We are concerned with the situation where $x_i$ is a scalar, i.e. $\mathbf{x}$ is a univariate time series. We also assume for simplicity that all series are the same length, $m$. Traditional time series machine learning tasks involve learning from a collection of $n$ time series,
$\mathbf{X}=\{\mathbf{x}_1,\dots,\mathbf{x}_n\}$.

The goal of TSCL is to take a collection of time series and group them together based on homogeny. Our focus is on the perennially popular $k$-means clustering algorithm~\cite{macqueen67kmeans} that forms representative centroids, or prototypes, for each cluster then assigns series to centroids based on a distance measure. We assume $k$ is known for all experiments (we outline how we identify $k$ in Section~\ref{sec:results}). $k$-means is a partitional clustering algorithm involving the following steps: 
\begin{enumerate}
    \item \textbf{initialisation}: create initial centroids (prototypes) for each cluster.
    \item \textbf{assignment}: assign each instance to the nearest centroid, based on a distance function.
    \item \textbf{update}: recalculate the centroids based on the new assignment.
    \item \textbf{stop or repeat}: determine whether to continue with a further iteration based on some stopping condition.
\end{enumerate}
Assignment and update are repeated in an expectation-maximisation style greedy algorithm that minimises the sum of squared error (SSE) objective function:

\begin{equation} SSE = \sum_{i=1}^{k} \sum_{j=1}^{n_i} d^2({\bf x}_{j}^{(i)}, {\bf c}_i)
\label{eq:sse}
\end{equation}
where $n_i$ is the number of time series assigned to cluster $i$, ${\bf x}_{j}^{(i)}$ is the $j^{th}$ time series assigned to $i^{th}$ cluster, ${\bf c}_i$ is the $i^{th}$ prototype/centroid time series, and $d$ is a distance measure between two series. For tabular clustering, $d$ is assumed to be the Euclidean distance for convergence reasons~\citep{bottou94convergence}. For TSCL, $d$ can be any time series specific distance. 

An important consideration when changing the distance measure is the method used to compute centroids. With the Euclidean distance, centroids are calculated as the arithmetic mean, as this minimises the SSE for that specific distance measure. However, when switching to a different distance measure, such as DTW, an alternative averaging technique that minimises the SSE for DTW must be applied to ensure proper convergence. We focus on two elastic distance functions, DTW and Move-Split-Merge~\cite{stefan13msm} and the possible averaging algorithms that could be used with them.

\subsection{Elastic distances}
\label{sec:elastic-distances}
\begin{figure}[!ht]
    \centering
    \begin{subfigure}[b]{0.45\textwidth}
        \centering
    \includegraphics[width=\textwidth]{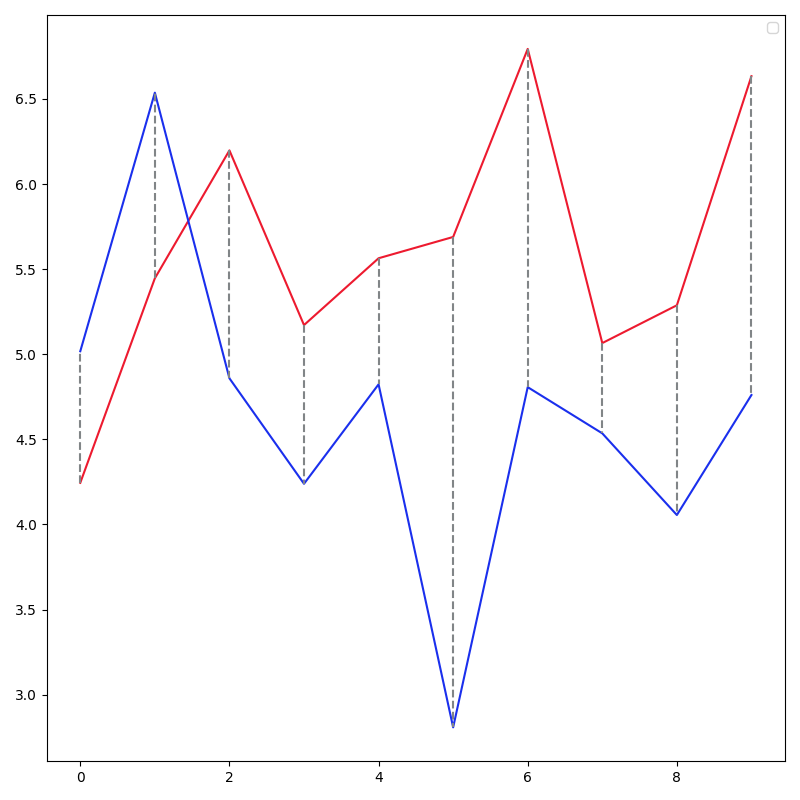}
    \caption{Euclidean alignment path.}
    \label{fig:point-to-point-alignment}
    \end{subfigure}
    \begin{subfigure}[b]{0.45\textwidth}
        \centering
        \includegraphics[width=\textwidth] {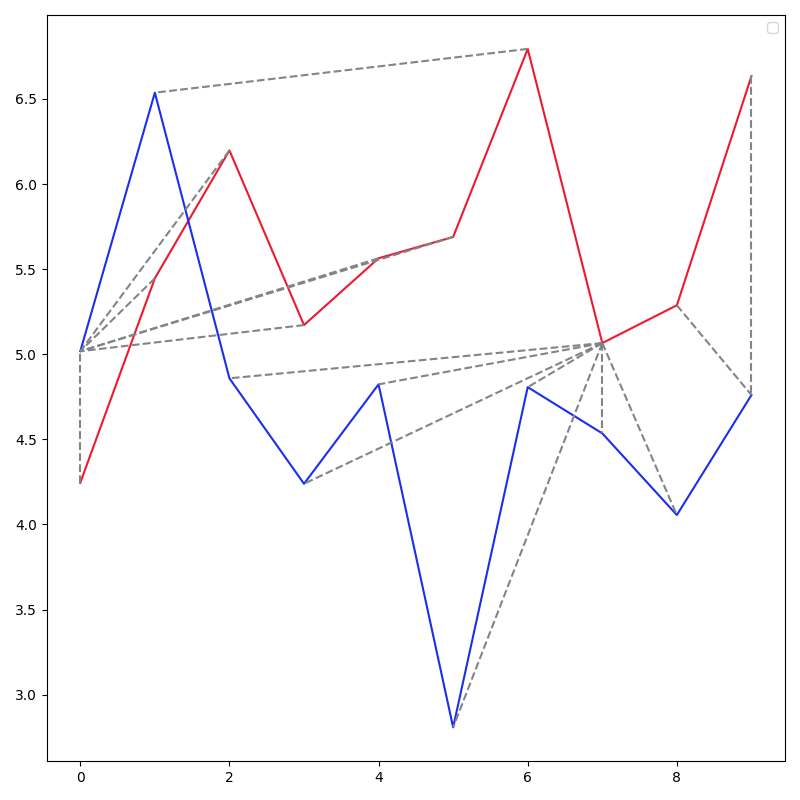}

        \caption{DTW alignment path.}
        \label{fig:dtw-alignment-no-bounding}
    \end{subfigure}
    \hfill
    \caption{Example of alignment between two time series when using the Euclidean distance and DTW distance. The dashed gray lines represents which points in the red time series are compared to in the blue time series.}
    \label{fig:alignment-path}
\end{figure}

Measuring the distance between time series is a primitive operation that can be used for a range of tasks such as classification, clustering, regression, anomaly detection and query retrieval. Distances that account for misalignment between two time series during computation are often called elastic distances. Figure~\ref{fig:point-to-point-alignment} gives and example of when $L_p$ measures such as Euclidean distance may not capture series similarity: a small offset between intrinsically similar series creates a large distance. Although many elastic distances have been proposed, we focus on two: DTW and MSM. Figure~\ref{fig:dtw-alignment-no-bounding} shows an example of using DTW to allign two series to better capture their similarity. DTW is the most popular elastic distance and the most researched for TSCL tasks, but MSM has been shown to be effective for $k$-means clustering~\cite{holder24review} and $k$-nn classification~\cite{lines15elastic}. MSM has the property of being a metric. We direct the reader to~\cite{holder24review, shifaz23elastic, abanda19distance} for comprehensive backgrounds into a wider range of elastic distances and~\cite{stefan13msm} for a proof of the metric property of MSM. 

\subsubsection*{Dynamic time warping (DTW)}
\label{sec:dtw}
Dynamic time warping (DTW) was first proposed for time series machine learning by~\cite{berndt94dtw} and has been used in thousands of publications since. DTW works by allowing one-to-many alignments (``warping'') of points between two time series~\cite{shifaz23elastic}. For time series classification, the one nearest neighbour (1-NN) classifier using a tuned version of DTW was considered state-of-the-art for many years and is still used as a baseline comparison for TSC benchmarks~\cite{middlehurst2024bakeoffredux}.

DTW uses dynamic programming to find the optimal path through a cost matrix ($CM$) that minimises the cumulative distance between two time series. It achieves this by constructing a pairwise matrix where each element is the cost of aligning a point from the first time series with a point from the second time series. The cost matrix for DTW is denoted as $CM_{dtw}$. Let $CM_{dtw}$ be an $(m+1) \times (m+1)$ cost matrix with indices starting from $i=0$ and $j=0$. The matrix $CM_{dtw}$ is initialised such that:

\begin{equation}
    \begin{aligned}
        CM_{dtw}(0,0) &= 0\\
        CM_{dtw}(i,0) &= +\infty, \quad 0 \leq i \leq m\\
        CM_{dtw}(0,j) &= +\infty, \quad 0 \leq j \leq m\\
    \end{aligned}
    \label{eq:cm-init}
\end{equation}
Once the $CM_{dtw}$ has been initialised the $CM$ is incrementally updated such that:
\begin{equation}
    \begin{aligned}
        CM_{dtw}(i,j) = (a_i - b_j)^2 + \min \begin{cases}
            CM_{dtw}(i-1, j-1) \\
            CM_{dtw}(i-1, j) \\
            CM_{dtw}(i, j-1)
            \end{cases}
    \end{aligned}
    \label{eq:dtw-cm}
\end{equation}
where $CM_{dtw}$ is the DTW $CM$, $i$ is a integer where $0 < i \leq m$, $j$ is a integer where $0 < j \leq  m$ and $a$ and $b$ are time series of length $m$. Once all values in $CM_{dtw}$ have been computed the DTW distance between time series $a$ and $b$ is given in Equation~\ref{eq:dtw}.
\begin{equation}
    d_{dtw}(a,b) = CM_{dtw}(m,m)
    \label{eq:dtw}
\end{equation}

A warping path (or alignment path) can also be extracted from the cost matrix. A warping path is defined as $P=<(e_1,f_1),(e_1,f_1),\ldots,(e_s,f_s)>$ where each value in $P$ is a pair of indices that define a traversal of matrix $CM$. A valid warping path must start at location $(1, 1)$, end at point $(m,m)$ and should not backtrack, i.e. $0 \leq e_{i+1}-e_{i} \leq 1$ and $0 \leq f_{i+1}- f_i \leq 1$
for all $1 \leq i < m$. An example of a warping path for DTW can be seen in Figure~\ref{fig:alignment-path}.

\subsubsection*{Move-Split-Merge (MSM)}
The alignment path for an elastic distance can be considered as a series of moves through a cost matrix. At any step an elastic distance algorithm can use one of three costs in forming an alignment: diagonal, horizontal or vertical. DTW assigns no explicit penalty for moving off the diagonal. Instead, it uses an implicit penalty (longer paths have more terms in the total distance). 

An alternative family of distance functions are based on the concept of edit distance. An edit distance, such as Move-Split-Merge (MSM)~\cite{stefan13msm}, considers a diagonal move as a match, a vertical move as an insertion and an horizontal move as a deletion. The move/match operation in MSM uses the absolute pointwise difference rather than the squared Euclidean distance used by DTW. The cost of the split operation is given by cost function $msm\_cost$ (Equation~\ref{eq:msm-cost}) with a call to $msm\_cost(a_i,a_{i-1},b_j, c)$. If the value being inserted, $b_j$, is between the two values $a_i$ and $a_{i-1}$ being split, the cost is a constant value $c$. If not, the cost is $c$ plus the minimum deviation from the furthest point $a_i$ and the previous point $a_{i-1}$ or $b_{j}$. The delete/merge is given by  $msm\_cost(b_j,a_{i},b_{j-1},c)$. Thus, the cost of splitting and merging values depends on the value itself and adjacent values. The  definition of MSM and the resulting distance are shown in Equations~\ref{eq:msm-cm} and~\ref{eq:msm} 
MSM satisfies triangular inequality and is a metric (see~\cite{stefan13msm} for a proof). We use a constant value of $c=1$ in all our experiments.

\begin{equation}
\begin{aligned}
msm\_cost(x,y,z,c) = \begin{cases}
    c &\textit{if }(y \leq x \leq z )\\
    c &\textit{if }(y \geq x \geq z )\\
    c+\min \begin{cases}
        |x-y| \\
        |x-z| 
    \end{cases} & \textit{otherwise}
\end{cases}
\end{aligned}
    \label{eq:msm-cost}
\end{equation}

\begin{equation}
    \begin{aligned}
        CM_{msm}(i,j) = \min\begin{cases}
                    CM_{msm}(i-1,j-1) + |a_i-b_j|\\
                    CM_{msm}(i-1, j) + msm\_cost(a_i, a_{i-1}, b_j, c)\\
                    CM_{msm}(i,j-1) + msm\_cost(b_j, a_i, b_{j-1}, c)

            \end{cases}
    \end{aligned}
    \label{eq:msm-cm}
\end{equation}

\begin{equation}
    d_{msm}(a,b) = CM_{msm}(m,m)
    \label{eq:msm}
\end{equation}
 $CM_{msm}$ is the $(m+1)\times(m+1)$ MSM cost matrix initialised to zeros, $i, j$ are integers where $0 < i, j \leq m$, $j$, $c$ is the cost for moving off the diagonal and $a$ and $b$ are time series of length $m$.

\begin{figure}[!ht]
    \centering
    \begin{subfigure}[b]{0.45\textwidth}
        \centering
        \includegraphics[width=\textwidth]
         {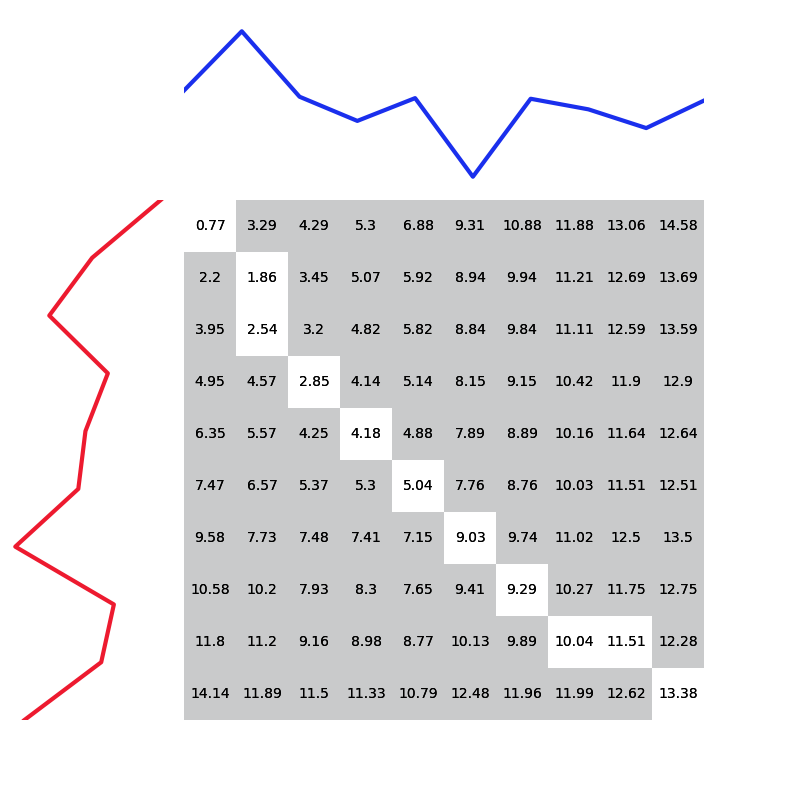}
        \caption{Optimal warping path (white squares) through $CM_{msm}$ between the red and blue time series where $c=1$.}
        \label{fig:msm-cm-visual}
    \end{subfigure}
    \hfill
    \begin{subfigure}[b]{0.45\textwidth}
        \centering
        \includegraphics[width=\textwidth]{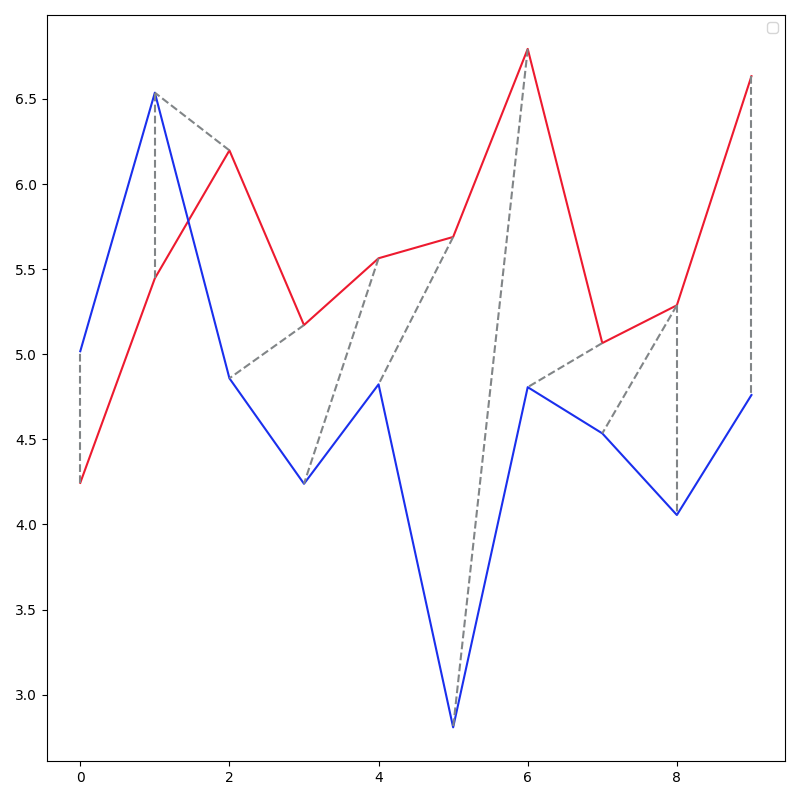}
        \caption{A visualisation of the MSM alignment between the red and blue time series where $c=1$.}
        \label{fig:msm-warping-path}
    \end{subfigure}
    \caption{Optimal MSM warping path through $CM_{msm}$ and a visualisation of MSM alignment between the two time series.}
    \label{fig:msm-visualisation}
\end{figure}

Figure~\ref{fig:msm-visualisation} illustrates how the MSM optimal alignment path is computed. Figure~\ref{fig:msm-cm-visual} presents the MSM cost matrix computed between the red and blue time series, with the white squares indicating the optimal warping path through the matrix. Figure~\ref{fig:msm-warping-path} depicts this MSM optimal warping path directly between the red and blue time series. 

\subsection{Time series averaging techniques}
\label{sec:time-series-averaging}
The objective of a time series averaging technique is to construct a time series that represents the centre of a collection of time series~\cite{schultz18ssgba}, often called a prototype (or in the context of $k$-means a centroid). The simplest way to compute the average of a collection of time series is by taking the arithmetic mean of all series over time points. This ignores the ordering, and for most time series machine learning tasks this will result in ignoring important features. Because of this, several time series-specific averaging techniques have been proposed to allow for offset between otherwise similar series.

\subsubsection{DTW Barycentre Average (DBA)}
\label{sec:dba}
DTW Barycentre Averaging (DBA)~\cite{petitjean11dba} is an averaging algorithm that considers the optimal alignment path between time series before an average is extracted. A barycentre refers to the central sequence that minimises the sum of squared distances between itself and a set of sequences~\cite{petitjean11dba}. When phrased as an optimisation problem, DBA aims to minimise the DTW Fréchet function \cite{frechet48frechetfunction}, i.e. if we have $p$ series in a collection, the problem is to find the centre ${\bf c}$ that minimises the objective:
\begin{equation}
\label{eq:fretchet}
    F({\bf X}) = \frac{1}{p} \sum_{i=1}^{p} d_{DTW}({\bf c}, {\bf x}_i)
\end{equation}
There is no closed solution to find the global minimum of the non-differentiable, non-convex Fréchet function~\cite{schultz18ssgba}. DBA uses a heuristic strategy to compute a new series that minimises the DTW distance to all members of the collection. A single pass of DBA  is described in Algorithm~\ref{algo:ba}. Each series is aligned with the current protoype (line 4), then the values warped into each position are collected (lines 5-6), before being averaged after all series in the collection are aligned (lines 7-8). 
\begin{algorithm}[htbp]
\caption{DBA(${\bf c}$,${\bf X}$)}
    \label{algo:ba}
\KwIn{
    
    ${\bf X}$: collection of $n$ time series\\
    ${\bf c}$: initial prototype\\
}
\KwOut{
    ${\bf c}$: new prototype
}

Let dtw\_path be a function that returns the a list of tuples that contain the indexes of the warping path between two time series.

Let $W$ be a list of empty lists, where $W_i$ stores the values in ${\bf X_p}$ of points warped onto centre point $c_i$.

    \For{$x \in {\bf |X|}$}{
	
        $P \leftarrow \text{dtw\_path}({\bf c},{\bf x})$
	    
            \For{$(i,j) \in P$}{
		    
                $W_{i} \leftarrow W_{i} \cup x_j$

            }
        }	
        \For{$i \leftarrow  1$ to $m$}{
	    
            $c_i \leftarrow mean(W_i)$

        }
        
    \Return $c$

\end{algorithm}
The heuristic DBA algorithm described in~\cite{petitjean11dba} begins with an initial average time series (input ${\bf c}$ for Algorithm~\ref{algo:ba}). This was randomly selected from the collection being averaged in the initial publication~\cite{petitjean11dba}. The initialisation was modified to be the medoid (using DTW) in the follow up~\cite{petijean16faster}. This requires finding all pairwise distances and is prohibitively expensive in run time. It was changed to the arithmetic mean in a subsequent version~\cite{petitjean17dbasynthetic}. Using DBA has been shown to significantly outperform the arithmetic mean for $k$-means clustering~\cite{petitjean11dba, schultz18ssgba, holder23mba, ali23shapedba}. DBA repeats Algorithm~\ref{algo:ba} until the objective function (Equation~\ref{eq:fretchet}) does not improve or until a maximum number of iterations is performed (set to 50 in all experiments).

\subsubsection{Elastic Barycentre Average}
\label{sec:elastic-barycentre-average}
The Elastic Barycentre Average~\cite{holder23mba} is a recently proposed generalisation of the DBA algorithm that can be applied with any elastic distance. While~\cite{holder23mba} demonstrated results specifically for the MSM elastic distance, they provided a generalised framework that allows for the use of any elastic distance that computes a complete alignment path through the cost matrix.
The Elastic Barycentre Average replaces the DTW alignment path with the alignment path of any other elastic distance in Algorithm~\ref{algo:ba}. It allows the clustering algorithm to easily leverage elastic distances that have been shown to be more effective for clustering than DTW~\cite{holder24review}. A recent variant of the Elastic Barycentre Average, Shape-DBA~\cite{ali23shapedba}, employs  shape-DTW~\cite{zhao18shapedtw} elastic distance measure to form barycentres.

\subsubsection{Stochastic Subgradient DTW Barycentre Average (SSG-DBA)}
\label{sec:ssg-dba}
The Stochastic Subgradient DTW Barycentre Average (SSG-DBA)~\cite{schultz18ssgba} is similar to DBA, but employs a Stochastic Subgradient (SSG) approach to try and optimise the Fréchet function rather than the greedy heuristic used in DBA. Both DBA and SSG-DBA are iterative methods that update a candidate solution through successive passes through the collection (epochs). The key difference is that DBA collates alignment paths to the centre then computes the average after the epoch whereas SSG-DBA updates the prototype after every alignment. The update is performed using a decaying learning rate. This means for a single pass or epoch through a collection of $p$ time series, DBA will perform $p$ alignment path computations and make $1$ update to the barycentre whereas SSG-DBA will perform $p$ alignment path computations and make $p$ updates to the barycentre on every epoch. Like DBA, SSG-DBA will continue updating in epochs until the stopping criteria of no improvement in the objective function (Equation~\ref{eq:fretchet}), or a maximum number of iterations is reached. Because it updates the prototype during the epoch, SSG-DBA requires a further $p$ distance calculations to test the stopping condition. This extra cost is justified through evidence of faster convergence than DBA~\cite{schultz18ssgba}.

\subsubsection{Soft-DTW Barycentre Average (soft-DBA)}
\label{sec:soft-dba}
Soft-DTW Barycentre Averaging (soft-DBA)~\cite{cuturi2017softdtw} is an averaging method that computes an exact minimum of the Fréchet function using soft-DTW. Unlike traditional elastic distances, soft-DTW is differentiable, enabling an exact solution that minimises soft-DTW across all time series in a collection. This contrasts with the Elastic Barycentre Average, which relies on non-differentiable elastic distances. Since the global minimum of the non-differentiable, non-convex Fréchet function is unknown~\cite{schultz18ssgba}, the Elastic Barycentre Average (and similar barycentre averaging techniques, such as SSG) can only provide estimates.

Soft-DBA with $k$-means has been shown to significantly outperform versions that use DBA, shape extraction~\cite{paparrizos16kshapes}, and shift-invariant averaging~\cite{jaewon11ksc}. However, this precision comes at high computational cost.

\subsubsection{Other time series averaging techniques}
\label{sec:other-averaging-techniques}
The four time series averaging functions described are the most relevant to this work. However, many other time series averaging techniques have been proposed. 
For TSCL tasks, the Shape Extraction~\cite{paparrizos16kshapes} average was developed for use with the $k$-Shape clusterer to minimise the cross-correlation distance Shape-Based Distance (SBD)~\cite{paparrizos16kshapes}. Additionally, the Scale-Shift Invariant average~\cite{jaewon11ksc} was proposed for use with the $k$-Spectral Centroid ($k$-SC) clusterer, minimising the Scale-Shift distance~\cite{jaewon11ksc}.

\subsection{TSCL Algorithms}
\label{sec:tscl-algorithms}
TSCL algorithms can be grouped into distance based algorithms, deep learning clusterers and feature based approaches. Our primary focus is distance based algorithms that use the $k$-means algorithm. A wide range of distance based $k$-means algorithms have been proposed. We focus on the following:

\begin{itemize}
    \item \textbf{$k$-means}~\cite{macqueen67kmeans} uses Lloyd's algorithm~\cite{lloyds82algo} with Euclidean distance assignment and arithmetic mean averaging. The algorithm is the same as used in traditional clustering and treats the time series as tabular data.
    \item \textbf{$k$-shape}~\cite{paparrizos16kshapes} is one of the most popular approaches that adapts Lloyd's with a novel distance and averaging method that is both scale and shift invariant. $k$-shape uses a shape-based distance measure (SBD) that utilises the cross-correlation of two time series.
    \item \textbf{$k$-spectral centroids ($k$-SC)}~\cite{jaewon11ksc} uses a scale-shift invariant distance in both assignment and averaging. The shift invariant distance is found by sliding the first series across the second using a wrap around and computing the Euclidean distance at each step. The scale-shift distance is the minimum over all windows.
    \item \textbf{$k$-DBA}~\cite{petitjean11dba} is another popular TSCL algorithm that adapts Lloyd's to use DTW distance in assignment and minimises the DTW SSE using the DBA averaging technique described in Section~\ref{sec:other-averaging-techniques}.
    \item \textbf{$k$-MBA}~\cite{holder23mba} uses Lloyd's algorithm with the MSM distance measure with the MSM SSE minimised using the MSM elastic barycentre average (MBA) described in Section~\ref{sec:elastic-barycentre-average}. 
    \item \textbf{$k$-shape-DBA}~\cite{ali23shapedba} uses Lloyd's algorithm with the shape-DTW distance measure with the shape-DTW SSE minimised using the shape-DTW elastic barycentre average (shape-DBA) described in Section~\ref{sec:elastic-barycentre-average}.
    \item \textbf{$k$-soft-DBA}~\cite{cuturi2017softdtw} uses Lloyd's algorithm with the soft-DTW distance measure and the soft-DTW is minimised with the soft-DBA averaging technique described in Section~\ref{sec:soft-dba}
    
\end{itemize}
Another popular partitional approach for TSCL is $k$-medoids~\cite{kaufman90pam}. $k$-medoids algorithms find a solution to the same optimisation problem as $k$-means but only considers training data instances as cluster prototypes.  The prototypes are called medoids rather than centroids. $k$-medoids algorithms are some of the easiest to adapt for TSCL as the distance measure can simply be swapped out from the traditional Euclidean distance to any other without concern of changing the algorithms objective function, since there is no averaging stage. ~\cite{holder23kmedoids} presented a through evaluation of $6$ different $k$-medoids algorithms including Alternate~\cite{lloyds82algo}, Partition Around Medoids (PAM)~\cite{kaufman90pam}, Clustering LARge Applications (CLARA)~\cite{kaufman90clara} and CLARA based on raNdomised Search (CLARANS)~\cite{ng02clarans} and found that PAM using MSM (\textbf{PAM-MSM})  elastic distance outperformed the current state-of-the-art TSCL approaches. 

An alternative approach to TSCL is to perform some form of transformation on the time series to form tabular data that can be used with traditional clustering algorithms, which we call feature based TSCL. A common way of doing this is to extract features or perform some form of dimensionality reduction. Some popular approaches include:

\begin{itemize}
    \item \textbf{Rocket Clustering (R-Clustering)}~\cite{jorge24rclustering} uses a modified version of the MiniROCKET~\cite{dempster21minirocket,dempster20rocket} feature extraction algorithm. Specifically R-Clustering adjusts the configuration of the hyperparameters to better suit clustering. Once the features have been created, the principal components of the features are then extracted  before being passed to a traditional $k$-means clusterer to produce the final clustering results.
    \item \textbf{Unsupervised Shapelets (U-Shapelets)}~\cite{zakaria12ushapelet} extracts shapelets from the data which are used to cluster the data. Shapelets are time series subsequences that are maximally representative of a class~\cite{ye09time}. For unsupervised tasks, unsupervised shapelets (u-shapelets) are extracted by sliding a window of a specified length over every time series in the dataset. Each subsequence is evaluated and ranked based on its utility, which indicates its discriminative power. The ranking process assesses how effectively a subsequence can separate subsets within the dataset, enabling the discovery of the most representative patterns without requiring class labels. Once the shapelets are extract the distance between each time series and each shapelets is computed. This distance map is treated as a feature vector for each time series which are then clustered using the traditional $k$-means algorithm.
    \item \textbf{Two-Step Time series Cluster (TTC)}~\cite{aghabozorgi15ttc} performs clustering in two steps. First time series are grouped according to similarity in time by applying an affinity search technique. Subsequently, for each cluster a prototype is defined according to the affinity of the time series belonging to it. The second step computes the DTW distances between the subclusters prototypes. The subclusters are merged based on similarity by means of the $k$-medoids standard clustering method to produce the final clustering.
\end{itemize}

There are numerous other TSCL approaches including deep learning based clustering algorithms e.g.~\cite{lafabregue22deep} and statistical model based approaches~\cite{caiado16clustering} and many more e.g. TADPole~\cite{begum16tadpoles}, SOMTimeS~\cite{javed24somtimes} and JET~\cite{wenig24JET}. These are beyond the scope of our study.

\section{The KASBA clustering algorithm}
\label{sec:kasba}

The $k$-means (K) accelerate (A) stochastic subgradient (S) Barycentre (B) Average (A) (KASBA) clusterer delivers competitive clustering performance comparable to the state-of-the-art TSCL algorithms, while reducing run time by up to three orders of magnitude. KASBA can work with any distance function, but is optimised to work with MSM~\cite{stefan13msm}, because it exploits the fact that MSM has the metric property.

We have implemented novel adaptations to the initialisation, assignment and update $k$-means components. One significant contribution spans all components: each stage requires a distance function, and KASBA consistently applies the same elastic distance function to each stage. It is not always clear whether this is done in related research~\cite{holder24tsclwithkmeans}. 

The stages of KASBA, described in Algorithm~\ref{algo:kasba}, include initialisation through adapting the k-means++ algorithm (line 2), update through a novel stochastic subgradient barycentre averaging approach to find centroids (line 5) and a fast assignment algorithm that exploits the triangle inequality of the distance (line 7) to greatly speed up the algorithm. These operations are described in detail in the subsequent sections. For clarity, we do not include all KASBA parameters in the algorithms. These are described in detail in Sections~\ref{sec:parameters} and~\ref{sec:results}. We have used standard default parameters and have not performed any tuning. One central component of KASBA is retaining information from each stage to optimise the next iteration. The centroids found in the initialisation (line 2) and in subsequent calls to recalculate\_centroids (line 6) are stored and used in subsequent calls to improve convergence. ${\bf p}$ contain minimum distances to centroids under labeling ${\bf l}$. These distances may change after centroid recalculation (line 6), and the data in ${\bf p'}$ can be used to make the assignment of cases in line 8 more efficient. The assignment returns the new cluster labels and the distances to the assigned centroids.

\begin{algorithm}[!ht]
    \LinesNumbered
    \caption{KASBA (${\bf X}$, $k$)}
    \label{algo:kasba}
    \KwIn{
    
    ${\bf X}$: collection of $n$ time series\\
    $k$: number of clusters\\
    }
    \KwOut{
    
        ${\bf C} = [{\bf c_1}, \ldots, {\bf c_k}]$: collection of $k$ centroids
        
        ${\bf l}= [l_1, \ldots, l_n]$ list of initial cluster labels of  length $n={\bf |X|}$

        ${\bf p}= [p_1, \ldots, p_n]$ list of distances to centroids

    }
\vspace{0.5cm}
    $d \leftarrow \text{setup\_distance}()$ \hspace{2cm}\textit{configure and return distance function}

    ${\bf C, l, p} \leftarrow \text{elastic\_kmeans\_pp}({\bf X}, k, d)$ \hspace{2cm}(Algorithm~\ref{algo:elastic-kmeans-plus-plus})

    $ continue \leftarrow$ {\bf true}
    
    $ iterations \leftarrow$ {\bf 1}

    \While{continue}{

        ${\bf C', p'} \leftarrow \text{recalculate\_centroids}({\bf X, C},{\bf l, p}, d, k)$ \hspace{0.1cm} (Algorithm~\ref{algo:recalc})

        \textit{${\bf C'}$ contains the new centroids,
        ${\bf p'}$ contains the distances to the new centroids}

        ${\bf l', p'} \leftarrow \text{fast\_assign}({\bf X, C'},{\bf l, p'},d,k)$ \hspace{0.73cm}(Algorithm~\ref{algo:fast_label})


        $iterations \leftarrow iterations + 1$

        \If{${\bf l} = {\bf l'} \lor iterations = max\_its$}{
            $continue \leftarrow$ {\bf false}
        }

        ${\bf l} \leftarrow {\bf l'}$

        ${\bf p} \leftarrow {\bf p'}$

        ${\bf C} \leftarrow {\bf C'}$
    
    }
    \Return ${\bf C,l, p}$
\end{algorithm}

\begin{algorithm}[!ht]
    \LinesNumbered
    \caption{elastic\_kmeans\_pp(${\bf X}$, $k$, $d$)}
    \label{algo:elastic-kmeans-plus-plus}
    \KwIn{

        ${\bf X}$: collection of $n$ time series
        
        $k$: number of clusters
        
        $d$: distance function for two series
        
    }

    \KwOut{
        
        ${\bf C} = [{\bf c_1}, \ldots, {\bf c_k}]$: collection of $k$ centroids
        
        ${\bf l}= [l_1, \ldots, l_n]$ list of initial cluster labels of  length $n={\bf |X|}$

        ${\bf p}= [p_1, \ldots, p_n]$ list of distances to centroids
    }
   \vspace{0.5cm}

    $a \leftarrow rand(|{\bf X}|)$ \hspace{0.5cm} {\em Select a random initial centroid index $a$ from $\{1, 2, \dots, |X|\}$}

    ${\bf c_1} \leftarrow {\bf X}_{a}$

    \For{$i \leftarrow 1$ \KwTo $|{\bf X}|$}
    {
        $p_i \leftarrow d({\bf x_i}, {\bf c_1})$
        
        $l_i \leftarrow 1$
    }
    \For{$i \leftarrow 2$ \KwTo $k$}
    {

        $s = sum(p)$
\hspace{0.5cm} {\em Find $m$, the  distribution for sampling the next centroid}
        
        \For{$j = 1$ \KwTo $|X|$}
        {
            $m_j \leftarrow p_j/s$
        }

        $a \sim m$ \hspace{0.5cm} {\em  Randomly select the next centroid index using distribution $m$}
        
        ${\bf c_{i}} \leftarrow {\bf X}_a$

        \For{$j = 1$ \KwTo $n$}
        {
            $t \leftarrow d({\bf x_j},{\bf c_i})$ 
            
            \If{$t < p_j$} 
            {
                    $l_j \leftarrow i$\hspace{0.5cm}  {\em Reassign to new centroid}
                    
                    $p_j \leftarrow t$
            }
        }
    }
    \Return ${\bf C,l, p}$
\end{algorithm}

\subsection{Elastic k-means++ initialisation}
\label{sec:elastic-kmeans++}
$k$-means++~\cite{arthur07kmeansplusplus} is one of the most popular and widely recommended methods for cluster initialisation in traditional clustering literature~\cite{calebi13initreview}. A commonly used alternative to $k$-means++ is to create random intitial centroids, either through Forgy initialisation (choose random cases)~\cite{forgy65forgyinit} or random initialisation (create synthetic random cases)~\cite{lloyds82algo}. This is used with a fixed number (often 10) restarts of the entire clustering algorithm, with the best clustering chosen based the restart that minimises some objective function by the most. For $k$-means this is the SSE (Equation~\ref{eq:sse}). In tabular clustering, $k$-means++ has been shown to outperform both Forgy and random initialisation with restarts, and is also significantly faster~\cite{calebi13initreview}. However, for TSCL random or Forgy initialisation with 10 restarts has been shown to outperform the traditional $k$-means++ using the Euclidean distance~\cite{holder24tsclwithkmeans}.  

Restarting $k$-means 10 times with different initial centres is computationally expensive. Moreover, the random placement of initial centroids significantly increases the time required for the algorithm to converge. This issue is even more pronounced in TSCL due to the added computational complexity of elastic distance functions. To address this, we propose the elastic $k$-means++ algorithm, which extends $k$-means++ to work with any elastic distance. Algorithm~\ref{algo:elastic-kmeans-plus-plus} provides a detailed outline of this approach.

In Section~\ref{sec:analysis}, we demonstrate that the elastic $k$-means++ algorithm produces significantly better initial centroids than $k$-means++ with Euclidean distance. Furthermore, it outperforms both the Forgy method and random initialisation with 10 restarts for TSCL. Notably, it also leads to significantly faster convergence. 

Algorithm~\ref{algo:elastic-kmeans-plus-plus} begins by randomly selecting the first prototype from the training data (line 1) and calculating distances to this initial centroid (lines 3–5). Subsequent prototypes are chosen based on a probability proportional to their distance from the nearest already-selected prototype (lines 7–9). In standard $k$-means++, Euclidean distance is used for these calculations. However, experiments have shown that for TSCL, using Euclidean distance performs significantly worse than employing restart schemes~\cite{holder24tsclwithkmeans}.

One of the key computational improvements we introduce for KASBA is the adaptation of $k$-means++ to elastic distances. Specifically, we incorporate the global end-to-end distance function into the $k$-means++ algorithm (lines 4 and 13). For clarity, minor optimisations are omitted from the pseudocode (lines 11–15). For instance, if a series has already been selected as a centroid, its distance to other centroids does not need to be recalculated. Additionally, the algorithm outputs the distances to the initial centroids, eliminating the need for recalculating these during the update stage.


\subsection{Update: recalculating centroids}

Finding centroids/prototypes with arithmetic means whilst assigning with an elastic distance creates poor clusterings~\cite{holder24review}. This is because using an averaging technique that does not minimise the SSE for the distance used can lead to unexpected convergence. DBA and similar barycentre averaging techniques (BA) have been proposed to minimise time series specific distance functions (see Section~\ref{sec:time-series-averaging}). However, whilst they significantly improve cluster quality compared to standard $k$-means, they are very slow. The BA algorithm is iterative and requires a distance calculation for all cluster members at each iteration. This can take a significant amount of time, particularly if there are a large number of instances. One approach to reducing the number of iterations is to use a search heuristic such as stochastic subgradient descent in the internal fitting (described in Section~\ref{sec:ssg-dba}) with the goal of reducing the overall number of iterations. 
 
\begin{algorithm}[ht!]
    \LinesNumbered
    \caption{recalculate\_centroids(${\bf X, C},{\bf l, p}, d, k$)}
    \label{algo:recalc}
    \KwIn{
    ${\bf X}$: collection of time series\\
    ${\bf C}$: current cluster centroids\\
    ${\bf l}$: current cluster labels\\
    ${\bf p}$: current cluster labels\\
    $d$: distance function\\
    $k$: number of clusters\\ 
    }
    \KwOut{
        ${\bf C}$: list of new cluster centroids\\
        ${\bf p}$: list of distances from series to currently assigned centroid 
    }
    \vspace{0.5cm}

    \For{$i \leftarrow 1$ \KwTo $k$}
    {
        ${\bf X_i} \leftarrow$ split\_by\_label(${\bf X,l},i)$\hspace{0.6cm}{\em find cluster $i$}

        $s \leftarrow $ sum\_by\_cluster(${\bf p}, i$) \hspace{1cm}{\em find current sum of distances to centroid}
        
        ${\bf c}_i,{\bf p}_i \leftarrow$ kasba\_average(${\bf X}_i, {\bf c}_i, d,s$) \hspace{0.4cm}(Algorithm~\ref{algo:elastic_ssg})

        ${\bf p} \leftarrow$ merge(${\bf p,p}_i$) \hspace{1.7cm}{\em overwrite old $p$ values whilst maintaining indices}
    
    }
    return ${\bf C, p}$    
\end{algorithm}

\begin{algorithm}[!ht]
    \LinesNumbered
    \caption{kasba\_average(${\bf X}, {\bf c}, d, dist$)}
    \label{algo:elastic_ssg}
    \KwIn{
    
    ${\bf X}$: collection of time series in one cluster\\
    ${\bf c}$: current centroid \\
    $d$: distance function\\
    $dist$: sum of distances to old centroid
    }
    \KwOut{
    
    ${\bf c}$: new centroid for ${\bf X}$\\  
    
    ${\bf p}$: distance from each series to new centroid ${\bf c}$    
    }
    \vspace{0.5cm}
    $r \leftarrow \min(|X|, \max(10, (s \times |X|)))$ \hspace{0.5cm}{\em Proportion to sample}
    
    $\eta \leftarrow 0.05$ \hspace{3.7cm}{\em Gradient descent learning rate}
    
    $max\_iters \leftarrow 50$

    \For{$i \leftarrow 1$ to $max\_iters$}{

        ${\bf X'} \leftarrow {\bf X}$ 

       \If{$i > 1$}{
       
        ${\bf X'} \leftarrow$ sample(${\bf X'},r)$ 
        }
    
        ${\bf c'} \leftarrow$ kasba\_update\_centroid(${\bf X'}, {\bf c}, d,\eta)$ (Algorithm ~\ref{algo:ssg_update})

        $dist' \leftarrow 0$

        ${\bf p'} = []$ \hspace{3.5cm}{\em New distances to centroid}
        
        \For{$j \leftarrow 1$ to ${\bf |X|}$}{

            $p'_j \leftarrow d({\bf c}, {\bf x}_j)$
            
            $dist' \leftarrow dist' + p_j$
        }

        \If{$dist \leq dist'$}{
            \Return ${\bf c, p}$ \hspace{0.5cm}{\em Current centroid no better than previous, return previous}
        }
        $dist \leftarrow dist'$ \hspace{1cm}{\em Update $dist, {\bf p}, \eta$ and ${\bf c}$ for the next iteration}
        
        ${\bf p} \leftarrow {\bf p'}$

        $\eta = 0.05 \cdot e^{- 0.1 \cdot i}$

        ${\bf c} \leftarrow {\bf c'}$
    }
    
    \Return ${\bf c , p}$
\end{algorithm}
Our update function recalculate\_centroids is described in Algorithm~\ref{algo:recalc}. This is called from line 6 of KASBA (Algorithm~\ref{algo:kasba})). recalculate\_centroids simply calls kasba\_average (Algorithm~\ref{algo:elastic_ssg}) on each current cluster. Algorithm~\ref{algo:elastic_ssg} is passed the current centroid ${\bf c}_i$. and the current sum of distances $s$. These are used to initialise the search. It returns the new centroid and ${\bf p'}$, the distance to the new centroid for the current assignment ${\bf l}$. These values are retained, overwriting the previous distances to the old centroids (line 5). This is important to optimise the assignment stage, described in Section~\ref{sec:label}. 

Function kasba\_average (Algorithm~\ref{algo:elastic_ssg}) is a barycentre stochastic gradient descent inspired by the version described in~\cite{schultz18ssgba}. It is adapted to work with any elastic distance, uses the previous centroid  and sum of distances as the initial starting point, introduces a subsampling randomisation to reduce the number of distance calls each epoch and uses a learning rate with exponential decay.

Algorithm~\ref{algo:elastic_ssg} is presented with the hard coded parameters we used in experimentation for clarity. On each iteration except the first, a proportion $s$ of the series are selected (line 5-7) and the centroid is updated with these series (line 8) using function kasba\_update\_centroid (Algorithm~\ref{algo:ssg_update}). The algorithm iterates until either the total distances to the centroid does not change (lines 11-15), or $max\_iters$ iterations are performed. Initialising the search with the centroid ${\bf c}$ from the previous round and the sum of distances to this centroid ($dist$) dramatically reduces the number of epochs. However, since $dist$ is calculated over all cluster members, we do not subsample on the first iteration to avoid premature convergence. We experimentally assess these changes in Section~\ref{sec:analysis}.

The batch update function kasba\_update\_centroid (Algorithm~\ref{algo:ssg_update}) aligns series in the subsample with the current centroid (line 2), then incrementally adjusts the centroid towards the alignments. It uses a learning rate (line 6) which is constant for the batch.

\begin{algorithm}[!ht]
    \LinesNumbered
    \caption{kasba\_update\_centroid( 
    ${\bf X}, {\bf c}, d, \eta$)}
    \label{algo:ssg_update}
\KwIn{\\
    ${\bf X}$: sample of cluster members\\
    ${\bf c}$: current centroid\\
    $\eta$: learning rate\\
    $d$: distance function \\
    }
\KwOut{${\bf c}$: Updated centroid}
\vspace{0.5cm}
    \For{${\bf x} \in {\bf X}$}{

        path $\leftarrow \text{find\_alignment\_path}({\bf c, x}, d$)  
        
        Let ${\Delta}$ be a zero array

        \For{each pair of indices $(j, k) \in path$}{
            $\Delta_k \leftarrow \Delta_k + (c_k - x_j)$
        }
        ${\bf c} \leftarrow {\bf c} - \eta \times \Delta$
    }
    
    \Return ${\bf c}$
\end{algorithm}

\subsection{Assignment of series to clusters}
\label{sec:label}

\begin{algorithm}[!ht]
    \LinesNumbered
    \caption{fast\_assign(${\bf X, C, l, p},d$)}
    \label{algo:fast_label}
    \KwIn{
        ${\bf X}$: collection of $n$ time series\\        
        ${\bf C}$: collection of $k$ centroids\\
        ${\bf l}$: previous cluster labels\\
        ${\bf p}$: distance to assigned centroid\\
        $d$: distance function\\
        
    }
    \KwOut{
         ${\bf l}$: list of new cluster labels
        ${\bf p}$: distance from series to their new assigned cluster\\
    }
    \vspace{0.2cm}
    ${\bf M} \leftarrow pairwise\_distances({\bf C})$
    
    \For{$i \leftarrow 1$ to $|{\bf X}|$}{

        $min\_dist \leftarrow  p_i$ \hspace{2cm}{\em Distance to current assignment for ${\bf x}_i$}
        
        $closest \leftarrow l_i$ 
        
        \For{$j \leftarrow 1$ to $k$}{

            \If{$j == closest$}
                {\bf continue}

            $ bound \leftarrow 2 \cdot M_{j,closest}$ \hspace{1cm}{\em Apply triangle inequality, try skip this distance}

            \If{$min\_dist < bound$} 
                {\bf continue} 
                
            $temp \leftarrow d({\bf x}_i,{\bf c}_j)$
            
            \If{$temp < min\_dist$}{

                $min\_dist \leftarrow temp$
                
                $closest \leftarrow j$

            }
        }
        $l_i \leftarrow closest$
        
        $p_i \leftarrow min\_dist$
        
    }
    \Return ${\bf l, p}$
\end{algorithm}

Once new centroids are found, the assignment stage involves calculating the distance from each series to all centroids. This requires $nk$ distance calculations and represents a significant proportion of the time taken when clustering with elastic distances. The distance function we use, MSM, has a property that DTW does not: it is a metric, and hence satisfies the triangle, equality, i.e.  if we have three series ${\bf x}$, ${\bf y}$ and ${\bf z}$, then 
$$d({\bf x},{\bf z}) \leq d({\bf x},{\bf y})+d({\bf y},{\bf z}).$$
This property means we can exploit ideas proposed in~\cite{elkan03elkankmeans} that use the triangle inequality to skip some distance calculations. 
 Suppose a series currently belongs to a cluster with centroid ${\bf c_a}$ and we want to see if it is closer to centroid ${\bf c_b}$. When the triangle inequality holds, it is shown in~\cite{elkan03elkankmeans} that 
$$\text{If } d({\bf c_a}, {\bf c_b}) \geq 2d({\bf x}, {\bf c_a}) \text{ then } d({\bf x}, {\bf c_b}) \geq d({\bf x}, {\bf c_a}).$$
This inequality states that if the distance between centroids is twice that between a series and its current centroid, then the distance between the series and the candidate must be greater than the distance to the current centroid. Informally, if a series could move to its current centroid and back in less distance that it takes to move centroid to centroid, it will be further to the new centroid and it can be discounted.

If we store the distance from a series to its centroid,  $d({\bf x},{\bf c_a})$ and have calculated the pairwise centroid distance function $d({\bf c_a},{\bf c_b})$ at the start of an assignment stage, we know that we will not need to move {\bf x} to a different cluster if the distance between the series and its current centroid is less than half the distance between centroids. We have stored the distance to the new centroids (${\bf p'}$) that are a side effect of the recalculation of centroids. The function fast\_assign (Algorithm~\ref{algo:fast_label}) exploits this to assign labels based on centroids. It starts by finding the pairwise distance between the new centroids (line 1). We show in Section~\ref{sec:analysis} that this dramatically reduces the number of distance calculations required despite introducing $k\cdot(k+1)/2$ new distance calls at the start.

\section{Experimental Set up}

\subsection{Datasets}
We conduct our experiments using the time series data from the University of California, Riverside (UCR) archive~\cite{dau19ucr}\footnote{\url{https://timeseriesclassification.com}}. Our focus is on univariate time series, and for all experiments, we utilise 112 of the 128 datasets available in the UCR archive. We exclude datasets that contain series of unequal length or missing values.

In supervised machine learning fields, such as TSC, datasets are typically divided into a training split and a test split. This approach helps prevent overfitting. However, in unsupervised tasks such as TSCL, labels are not provided to the learning algorithm. This means that overfitting is not a concern, and any patterns or structures learned by the model must come from the model's inherent understanding of the training data itself, rather than from predefined labels~\cite{javed20benchmark}. Consequently, a common approach in clustering is to provide the model with all available data during training, since a separate prediction step is not required.

However, there are problems with comparing clustering algorithms on data used in training. Often, clustering forms an unsupervised component of a supervised pipeline for tasks such as channel selection~\cite{dhariyal23channel}. Assessing clusterers used in supervised pipelines on training data may lead to a biased view of performance on unseen data. We follow the recent trend in TSCL~\cite{lafabregue22endtoenddeeplearning,holder24review} and evaluate on the unseen predefined test split in the UCR archive. We use the labels only in the cluster evaluation. One such evaluation is to understand how well clusterers can extrapolate meaningful general patterns from the data. A model that performs well on unseen data has probably learned more  about the underlying properties of the data, potentially making its clusterings more valuable. As such our primary form of evaluation will focus on training a clusterer on the predefined train split for the UCR archive, followed by evaluation using the predefined test split. However, we acknowledge that a large body of the TSCL literature evaluate by combining the training and test splits into one large dataset for reasons described above. Hence, we also provide results and evaluation for KASBA using the combined training and test split. We believe this approach showcases clusterers utility for many more use cases and improves the evaluation of TSCL models.

Finally, we z-normalise all datasets prior to clustering. We are aware that some argue ``\textit{any improvement resulting from pre-processing (normalisation) should not be attributed to the clustering method itself}''~\cite{javed20benchmark}, others contend that ``\textit{in order to make meaningful comparisons between two time series, both must be normalised}''~\cite{rakthanmanon13trillionsubsequence}. Therefore, the decision to normalise or not remains an ongoing research question. Ideally, with unlimited time and computational resources, we would run both normalised and un-normalised experiments to compare the results. However, this is not feasible given our constraints, and we are forced to make a choice. Following the recommendation of~\cite{keogh03benchmark} and~\cite{rakthanmanon13trillionsubsequence} we choose to normalise our data. 

All of our experiments were run on a single thread of an Ice Lake Intel Xeon Platinum 8358 2.6GHz processor on a shared high-performance computer (HPC) cluster with a seven day computational runtime limit. Hence, we only report results for algorithms that complete in seven days.

\subsection{Clusterer configuration}
\label{sec:parameters}
All the clusterers we consider in this paper use the parameter $k$ which defines the number of clusters that should be formed. For simplicity we set a value that is equal to the number of unique class labels defined by the UCR archive for a given dataset. $k$-means assumes the number of clusters $k$ is set a priori. There are a range of methods of finding $k$. These often involve iteratively increasing the number of clusters until some stopping condition is met. This can involve some form of elbow finding or unsupervised quality measure, such as the silhouette value~\citep{rousseeuw87silhouettecoefficient}.  

\subsubsection*{Evaluation}

We measure performance on the 112 UCR datasets using four common clustering performance measures. \textbf{Clustering accuracy (CLACC)} is the number of correct predictions divided by the total number of cases. To determine whether a cluster prediction is correct, each cluster has to be assigned to its best matching class value. This can be done naively, taking the maximum accuracy from every permutation of cluster and class value assignment $S_k$ or more efficiently using a  combinatorial optimization algorithm.
$$
    CLACC(y,\hat{y}) = \max_{s \in S_k} \frac{1}{|y|} \sum_{i=1}^{|y|}
    \begin{cases}
        1, & y_i = s(\hat{y}_i) \\
        0, & \text{otherwise}
    \end{cases}
    \label{eqn:accuracy}
$$
The rand index measures the similarity between two clustering by comparing predictions of pairs of instances with the ground truth. If the predicted values are the same for a pair and the ground truth of the pair is also the same, this is a positive count for the rand index. The rand index is popular and simple. However, it is not comparable across problems with different number of clusters. The \textbf{adjusted rand index (ARI)} compensates for this by adjusting the RI based on the expected scores on a purely random model. 

The mutual information is a function that measures the agreement of a clustering and a true labelling, based on entropy. \textbf{Normalised mutual information (NMI)} rescales mutual information onto $[0,1]$, and \textbf{adjusted mutual information (AMI)} adjusts the MI to account for the class distribution.

We use a range of tools to summarise and compare multiple clusterers on multiple datasets. We compare ranks using an adaptation of the critical difference (CD) diagram~\citep{demsar06comparisons}, replacing the post-hoc Nemenyi test with a pairwise comparison using the Wilcoxon signed-rank tests, and cliques formed using the Holm correction recommended by~\cite{garcia08pairwise} and~\cite{benavoli16pairwise}. We use $\alpha=0.05$ for all hypothesis tests. Critical difference diagrams such as those shown in Figure~\ref{fig:baseline} display the average ranks of estimators and cliques of clusterers as solid lines. A clique represents a group where there is no significant different between clusterers. CD diagrams are useful, but when presented alone they can mask similarities. We also present summary performance statistics in a heat map proposed in~\cite{ismail2023approach} and use scatter plots for pairwise comparison.

\subsection{Reproducibility}
We conducted all experiments using the \texttt{aeon} open source time series machine learning toolkit\footnote{\url{https://github.com/aeon-toolkit/aeon}}. All results were evaluated with the \texttt{tsml-eval} open source package. Additionally, we have provided a notebook that outlines the steps required to run and reproduce our results\footnote{\url{https://github.com/time-series-machine-learning/tsml-eval/tree/main/tsml_eval/publications/clustering/kasba/kasba.ipynb}}. Furthermore, we have included all the results reported in this paper as separate CSV files for each evaluation metric alongside the notebook.

\section{Results}
\label{sec:results}
We organise our results into three sections to enhance clarity. Section~\ref{sec:baseline} compares KASBA with closely related benchmark algorithms. Section~\ref{sec:msm-results} evaluates KASBA against its variants without optimisations and related medoid-based approaches. Lastly, Section~\ref{sec:others} examines KASBA’s performance relative to a broader range of TSCL algorithms.

For our experiments, we use the clusterers described in Section~\ref{sec:tscl-algorithms}. All the clusterers assume the value of $k$ is known beforehand and set to the true number of clusters.
In all cases, we use the clusterer parameters recommended in the literature. Table~\ref{tab:distance-model-configuration} summarises the configuration of various distance-based clusterers. 

\begin{table}[!ht]
    \vspace{0.25cm}
    \centering
    \makebox[\textwidth][c]{%
    \begin{tabular}{|l|l|l|l|l|l|l|}
        \hline
         Algorithm & Initialisation  & Distance & Assignment  & Centroid  & Max Its &  Parameters\\
        \hline
        \textbf{KASBA} & $k$-means++ MSM & MSM & KASBA assign & KASBA  average & 300 & $c=1.0$ \\
        \textbf{DBA} & $k$-means++ DTW & DTW & Lloyd's assign & DBA & 300 & - \\
        \textbf{shape-DBA} & $k$-means++ shape-DTW & shape-DTW & Lloyd's assign & shape-DBA & 300 & $reach=15$ \\
        \textbf{soft-DBA} & $k$-means++ soft-DTW & soft-DTW & Lloyd's assign & soft-DBA & 300 & $\gamma=1.0$  \\
        \textbf{MBA} & $k$-means++ MSM & MSM & Lloyd's assign & MBA & 300 & $c=1.0$  \\
        \textbf{Euclid} & Forgy $10$ Restarts & Euclidean & Lloyd's assign & Arithmetic mean & 50 & - \\
        \textbf{MSM} & Forgy $10$ Restarts & MSM & Lloyd's assign & Arithmetic mean & 50 & $c=1.0$  \\
        \textbf{$k$-Shape} & Forgy $10$ Restarts & SBD & Lloyd's assign & Shape extraction & 50 & - \\
        \textbf{$k$-SC} & Forgy $10$ Restarts & $k$-SC dist & Lloyd's assign & $k$-SC average & 50 & $max\_shift=m$ \\
        \textbf{PAM-MSM} & Forgy $10$ Restarts & MSM & PAM-SWAP & MSM medoids & 50 & $c=1.0$ \\
        \hline
    \end{tabular}%
    }
    \vspace{0.25cm}
    \caption{Distance-based clustering algorithm configuration.}
    \label{tab:distance-model-configuration}
\end{table}

We are missing some results for specific datasets for certain clusterers. In this case either the algorithm’s run time exceeded the $7$-day limit or the algorithm repeatedly formed empty clusters and failed to converge. We explicitly indicate the number of datasets used in each set of results. A complete list of the missing results, along with the corresponding reasons, is presented in Tables~\ref{tab:test-train-split-missing-datasets} and~\ref{tab:combined-test-train-split-missing-datasets}.

\subsection{KASBA against distance based k-means algorithms}
\label{sec:baseline}

As discussed in Section~\ref{sec:tscl-algorithms}, $k$-means addresses an optimisation problem that seeks to minimise the SSE. The most widely used approach to this problem is Lloyd’s algorithm. For TSCL, Lloyd’s algorithm remains the standard, though it is adapted by modifying the distance measure and averaging technique. For this section we use $7$ other clusterers: DBA, shape-DBA, soft-DBA, Euclid, MSM, $k$-Shape and $k$-SC. The parameters used are presented in Table~\ref{tab:distance-model-configuration}.

For Euclid, MSM, $k$-Shape and $k$-SC we use Forgy initialisation with 10 restarts, returning the clustering with the lowest inertia (SSE) as the final result. In contrast, DBA, shape-DBA and soft-DBA and KASBA employ the elastic $k$-means++ with DTW, shape-DTW, soft-DTW and MSM respectively. For clusterers compatible with the elastic $k$-means++ algorithm proposed in Section~\ref{sec:elastic-kmeans++}, we have chosen to apply it to maintain consistency in our results.

\begin{figure}[htb]
    \centering
            \begin{tabular}{c c}
    \includegraphics[width=0.5\linewidth]{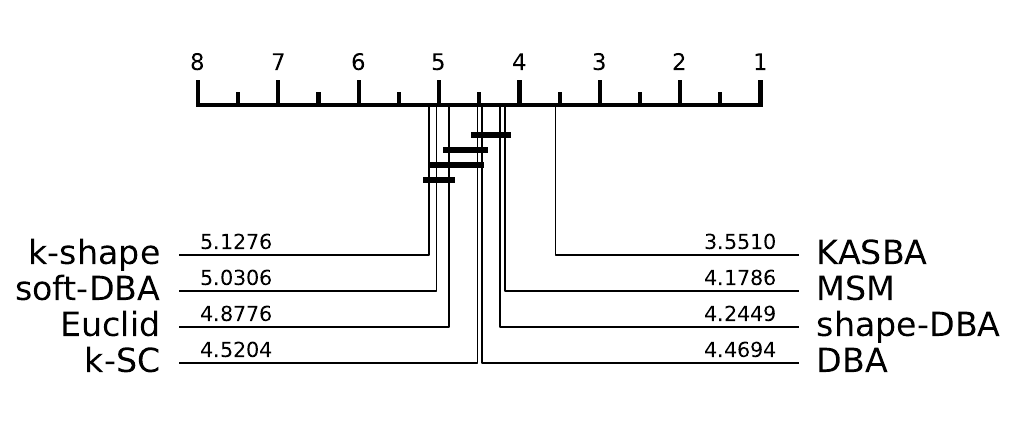} &
    \includegraphics[width=0.5\linewidth]
    {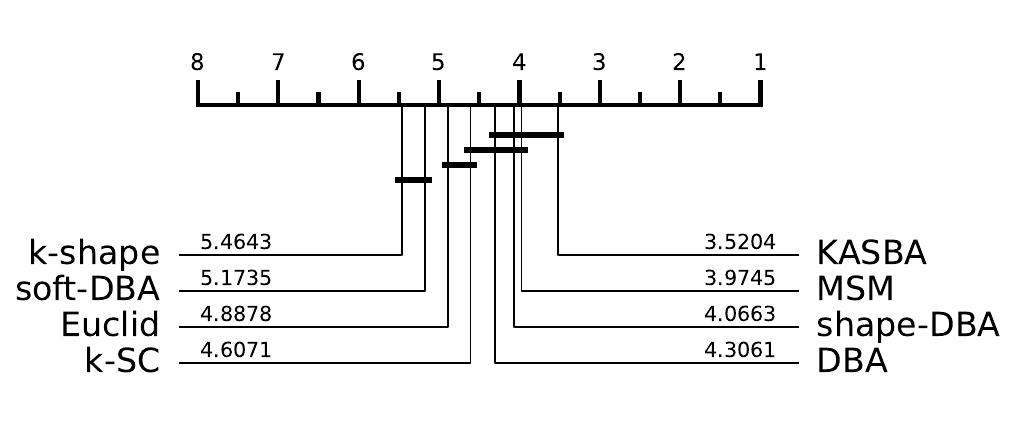}\\
    Accuracy & Adjusted Rand Index \\
    \includegraphics[width=0.5\linewidth]
    {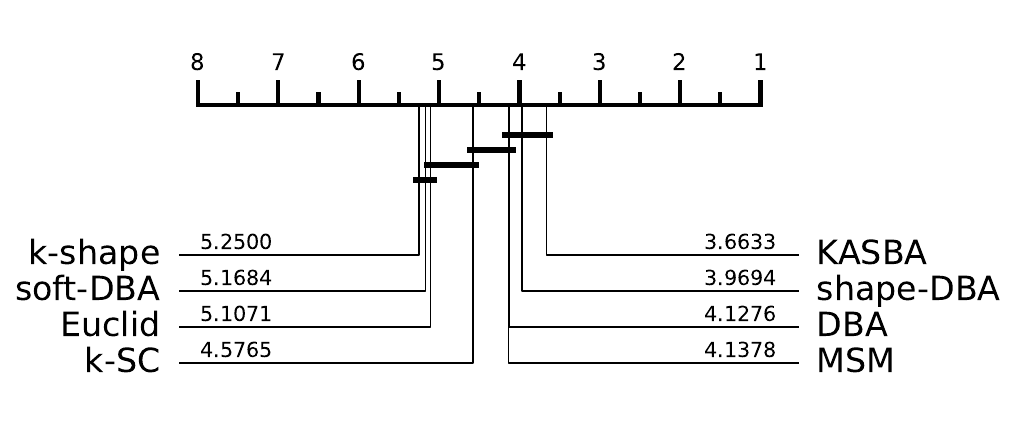} & 
    \includegraphics[width=0.5\linewidth]
    {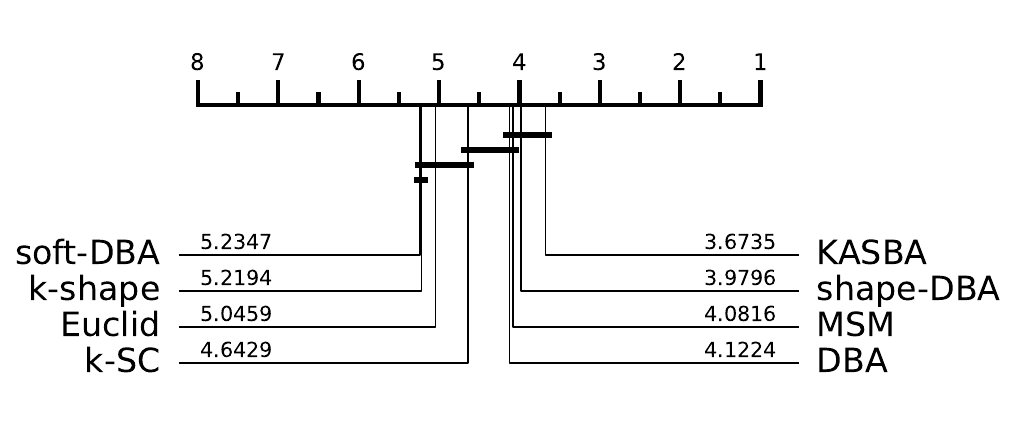} \\
    Average Mutual Information & Normalised Mutual Information \\
    \end{tabular}
    \caption{KASBA against seven benchmark algorithms on test data, averaged over the 98 UCR archive data completed by all algorithms using the default train-test split.}
    \label{fig:baseline}
\end{figure}

Figure~\ref{fig:baseline} shows the average ranks of KASBA against these seven $k$-means based clusterers using the default train/test split. KASBA is the top ranked algorithm for all four evaluation metrics and is significantly better than the others for accuracy. MSM, Shape-DBA and DBA are consistently the best of the others.

\begin{figure}[htb]
    \centering
            \begin{tabular}{c c}
    \includegraphics[width=0.5\linewidth]{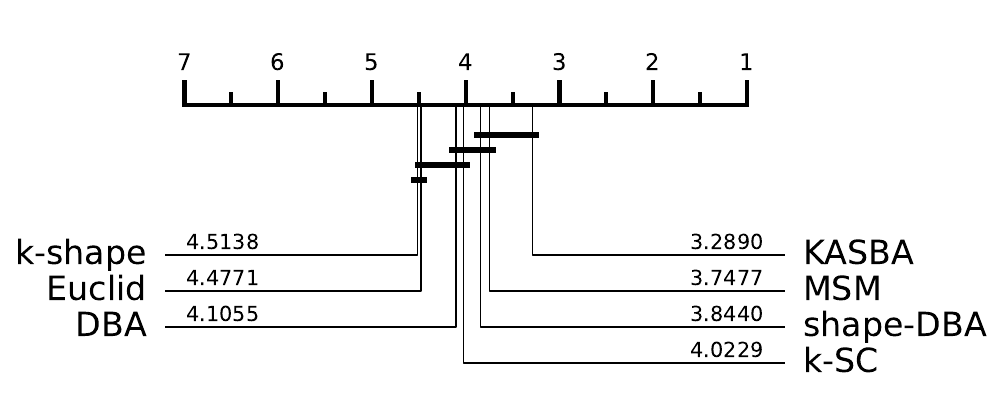} &
    \includegraphics[width=0.5\linewidth]
    {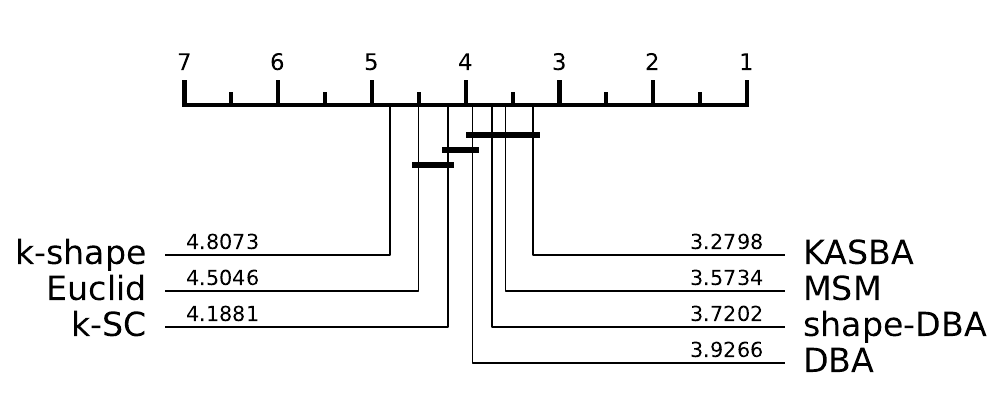} \\
    Accuracy & Adjusted Rand Index \\
    \includegraphics[width=0.5\linewidth]
    {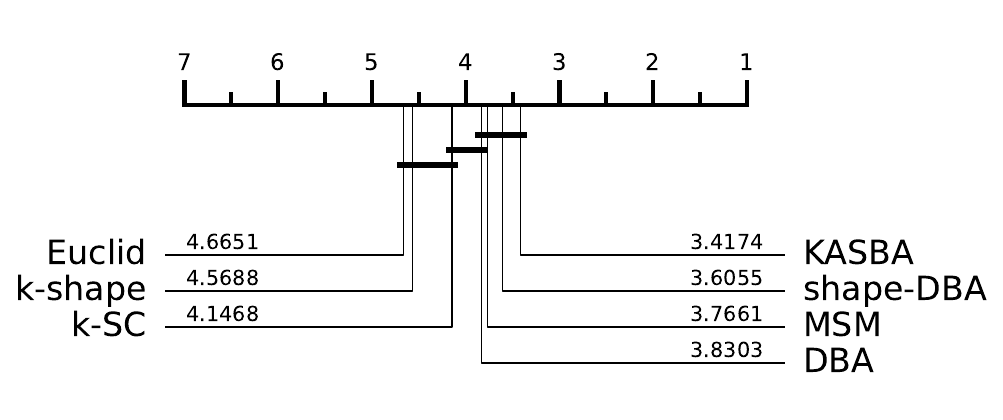} & 
    \includegraphics[width=0.5\linewidth]
    {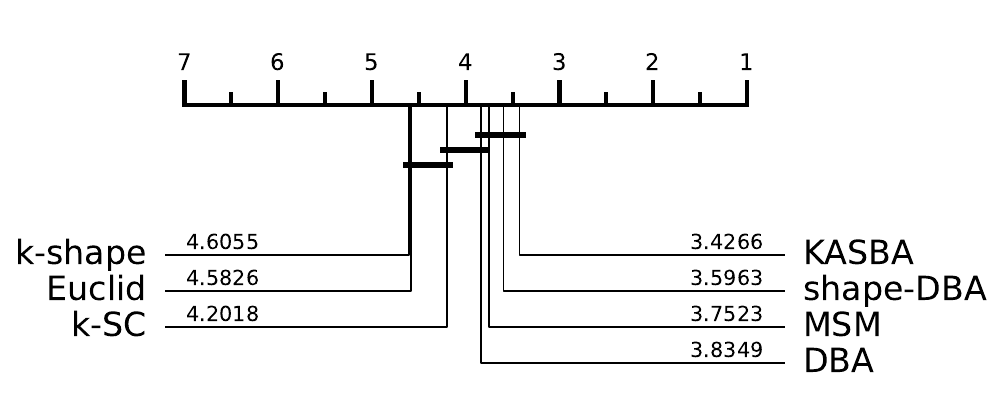} \\
    Average Mutual Information & Normalised Mutual Information \\
    \end{tabular}
    \caption{KASBA against seven benchmark algorithms on test data, averaged over the 109 UCR archive data completed by all algorithms using the default train-test split.}
    \label{fig:baseline-no-soft-dba}
\end{figure}

\begin{figure}[htb]
    \centering
    \includegraphics[width=1.0\linewidth]{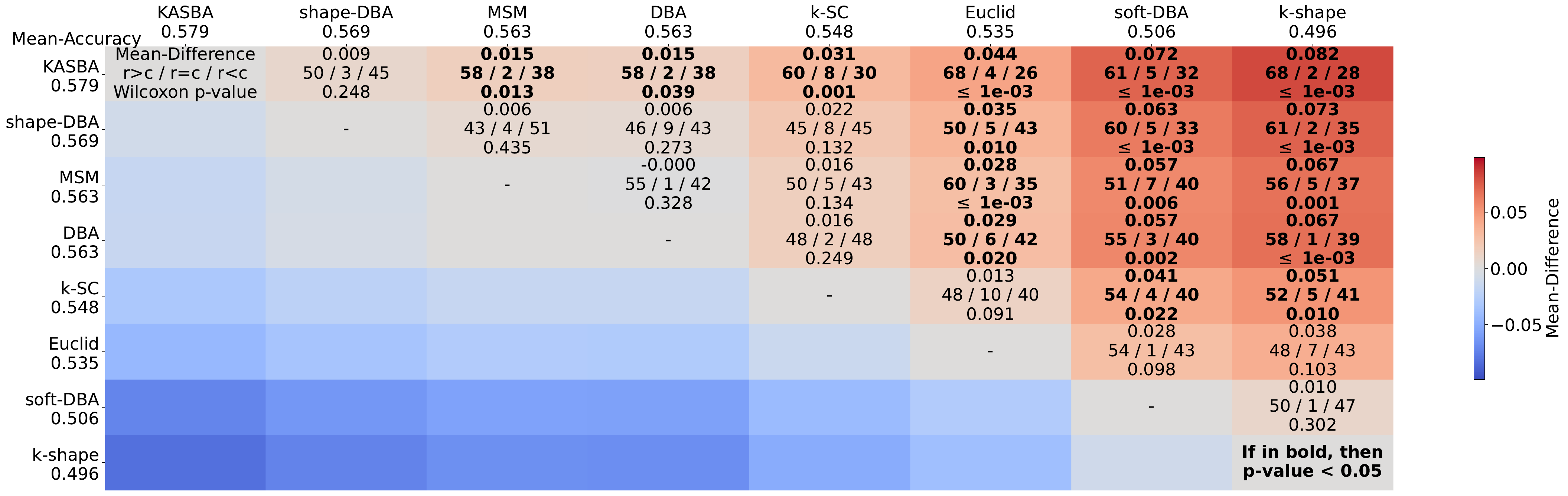}
    \caption{Summary performance measures for eight clustering algorithms for clustering accuracy, including mean difference (top), wins/ties/losses (middle) and p-value for a one sided Wilcoxon sign ranked test (unadjusted for multiple testing, bottom).}
    \label{fig:heatmap}
\end{figure}

Critical difference diagrams can mask relative performance of estimators due to the linear construction of cliques. For example, KASBA is significantly better than MSM and Shape-DBA for accuracy and so is not in the same clique. In addition, Figure~\ref{fig:baseline} would suggest that MSM is better (though not significantly) than Shape-DBA. However, the relative performance in terms of clustering accuracy summarised in the heat map in Figure~\ref{fig:heatmap} (described in~\cite{ismail2023approach}) shows there is no significant difference between KASBA and Shape-DBA.

\begin{figure}[htb]
    \centering
            \begin{tabular}{c c}
    \includegraphics[width=0.5\linewidth]
    {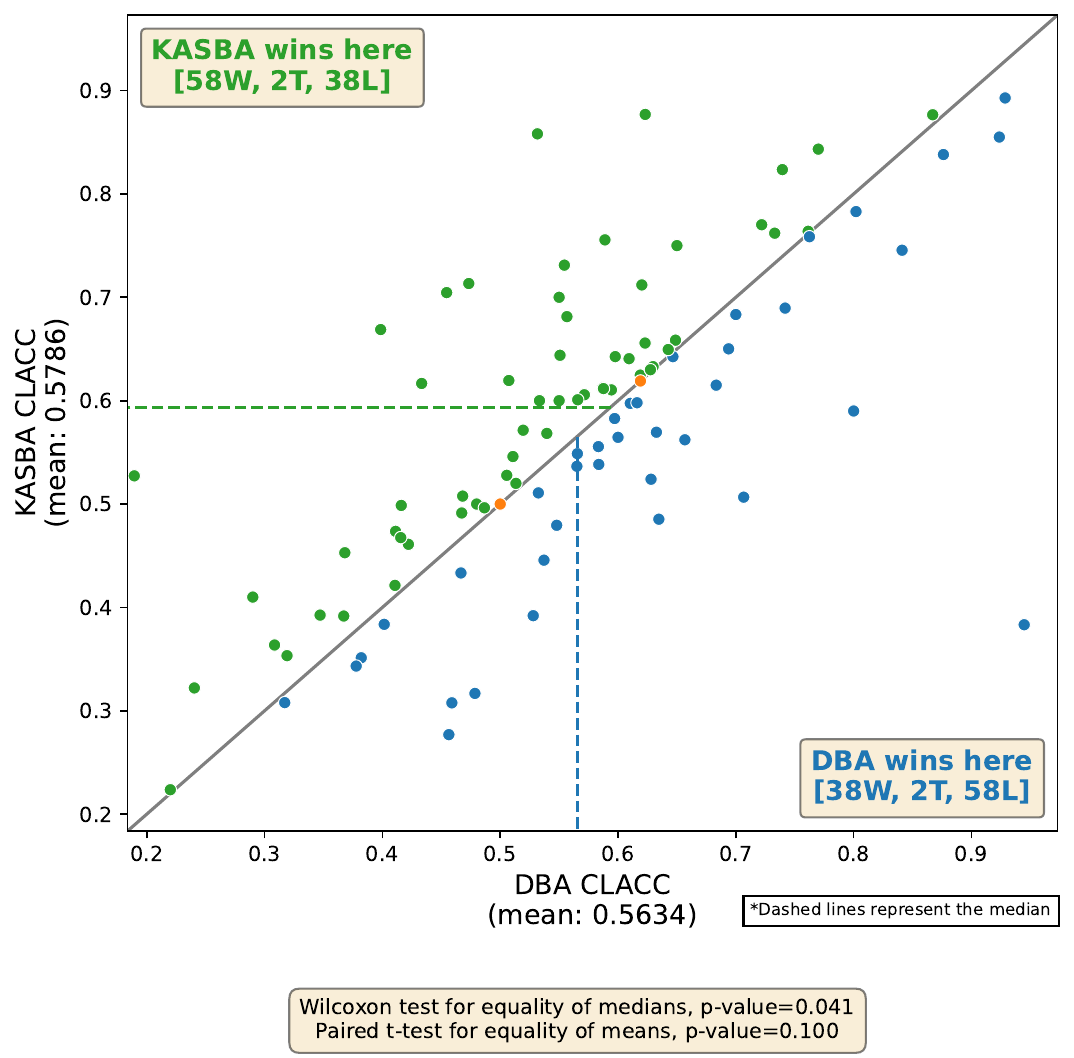} &
    \includegraphics[width=0.5\linewidth]
    {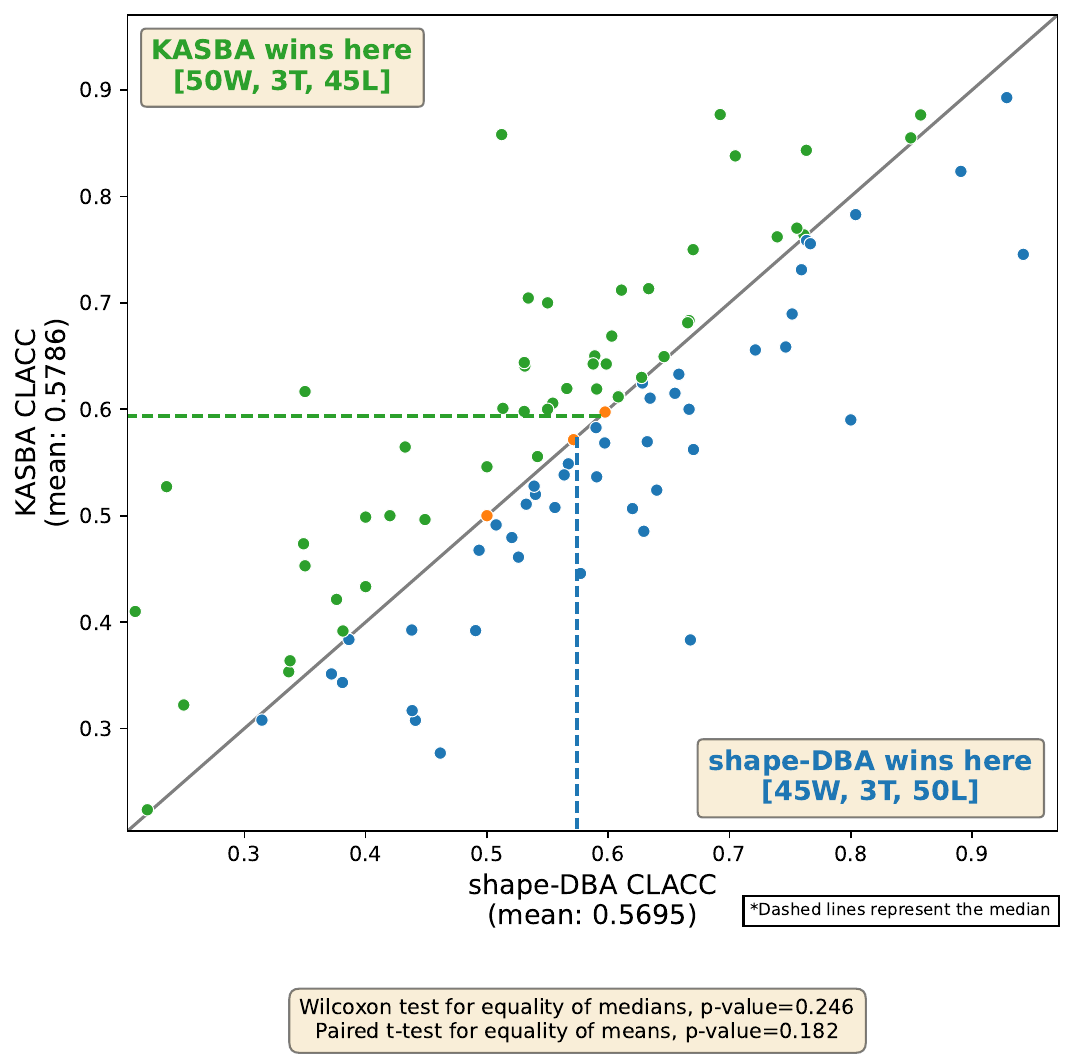}\\
    Accuracy & Adjusted Rand Index \\
    \end{tabular}
    \caption{Scatter plots of KASBA against DBA and Shape-DBA for classifier accuracy.}
    \label{fig:scatter}
\end{figure}

Figure~\ref{fig:heatmap} indicates that when we order by average values rather than average ranks, Shape-DBA is ranked second, and MSM third. This implies that MSM does well on ranks, but its averages are bought down by occasional very bad clusterings. Figure~\ref{fig:scatter} shows the scatter plots for accuracy KASBA against DBA and Shape-DBA.
KASBA is the top ranked algorithm. It is significantly better than the commonly used benchmarks $k$-Shape, soft-DBA and $k$-SC. It is better than MSM, DBA and Shape-DBA, but the improvement is relatively small. Where KASBA really excels is in run time. Table~\ref{tab:times} summarises the run times for completing 98 datasets for the train/test experiments. 
\begin{table}[!ht]
    \vspace{0.25cm}
    \centering
    \begin{tabular}{|l|l|l|l|l|}
        \hline
Estimator  & Median (mins) & Mean (mins) & Max (hrs) & Total (hrs) \\ \hline
Euclid     & 0.0           & 0.0         & 0.0       & 0.1         \\
KASBA & 0.0        & 0.5       & 0.3     & 0.8       \\
k-Shape    & 0.4           & 1.3         & 0.5       & 2.2         \\
DBA        & 0.9           & 24.3        & 10.7      & 39.6        \\
k-SC       & 1.8           & 28.8        & 11.3      & 47.0        \\
MSM        & 0.4           & 93.7        & 136.6     & 153.1       \\
Shape-DBA  & 1.7           & 128.2       & 48.1      & 209.4       \\
Soft-DBA   & 17.2          & 870.4       & 165.5     & 1421.7      \\ \hline
    \end{tabular}
    \vspace{0.25cm}
    \caption{Summary of training times for eight clusterers on 98 UCR datasets completed. Timings are for a sequential training runs for all 98 datasets.}
    \label{tab:times}
\end{table}

\begin{figure}[htb]
    \centering
            \begin{tabular}{c c}
    \includegraphics[width=0.5\linewidth]{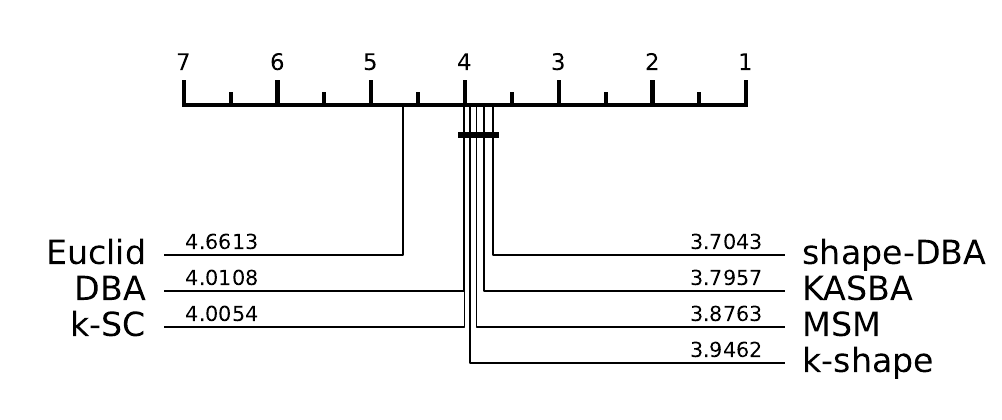} &
    \includegraphics[width=0.5\linewidth]
    {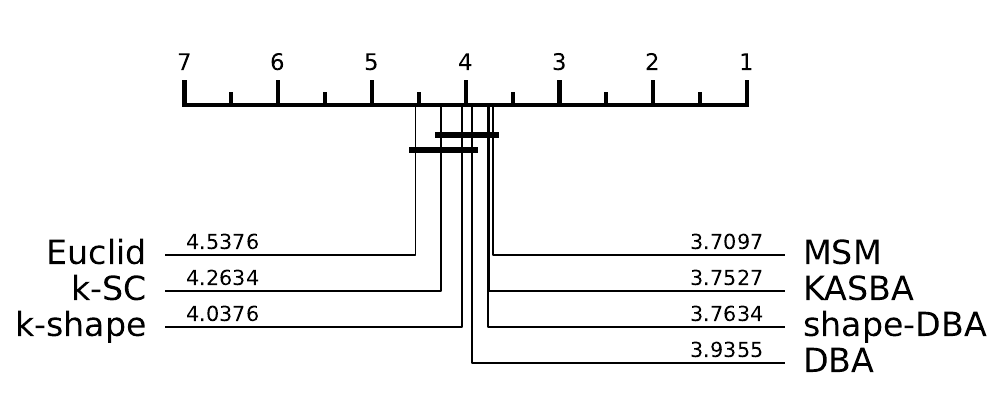}\\
    Accuracy & Adjusted Rand Index \\
    \includegraphics[width=0.5\linewidth]
    {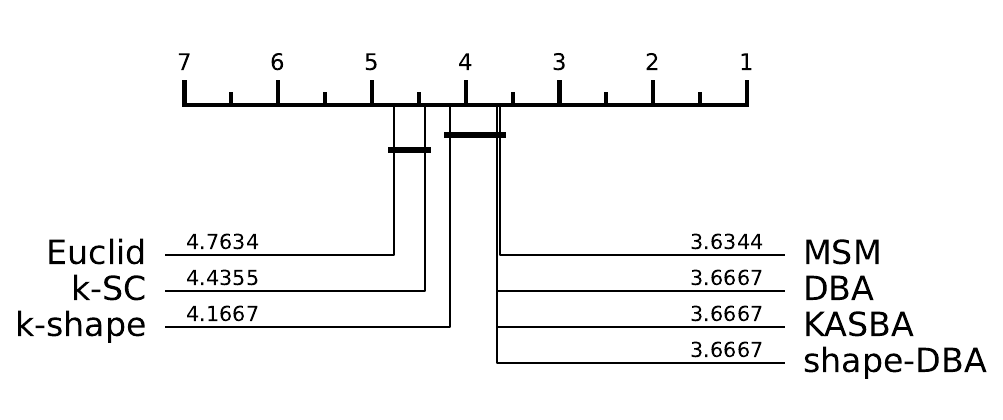} & 
    \includegraphics[width=0.5\linewidth]
    {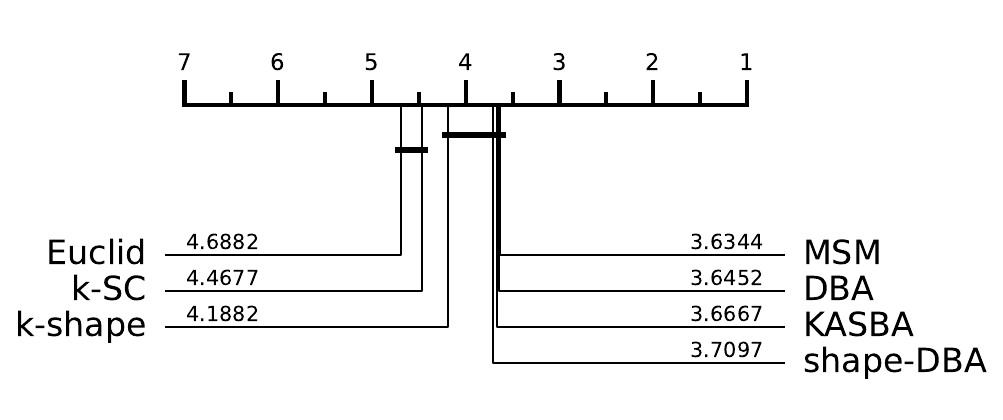} \\
    Average Mutual Information & Normalised Mutual Information \\
    \end{tabular}
    \caption{KASBA against seven benchmark algorithms using train set performance on combined data, averaged over the 93 UCR archive data completed by all algorithms.}
    \label{fig:combined}
\end{figure}

KASBA completes the 98 problems in just over 45 minutes, whereas Shape-DBA takes over eight days and Soft-DBA would take nearly two months (59 days). KASBA is around fifteen times slower than standard $k$-means with Euclidean distance but performs significantly better. KASBA is one to three orders of magnitude faster than all but the worst performing algorithm on the train/test data ($k$-Shape). All algorithms show a skew between mean and median. This is due to the larger number of small problems in the archive. However, the ratio of mean to median identifies a problem with some of these algorithms failing to converge in a reasonable time. The mean for KASBA is about 2.5 times the median. The mean for MSM is approximately 200 times the median. The single slowest problem for MSM is PigCVP. This is a strange dataset: it has 50 classes and only 100 training instances. Clearly, this is unusual and MSM is running for its maximum iterations. Conversely, KASBA converges in 10 minutes. This highlights how robust KASBA is. Shape-DBA does well in terms of performance, but is slow and also has convergence issues. Soft-DBA performs poorly in terms of performance and run time. This is surprising. We also ran it with 10 restarts and it did better, but it was so slow to make it impractical. $k$-Shape takes over double the amount of time to fit compared to KASBA, and Figure~\ref{fig:baseline} shows that $k$-Shape finds poor clusters when used with a train/test split. We repeated our experiment by combining the training and test splits. We then perform the same experiment evaluating the algorithms on the combined data. Figure~\ref{fig:combined} shows the results for our four performance statistics. There is now no significant difference between the top  five algorithms.

We believe the differences are simply harder to detect using the combined data, and that it gives a potentially skewed view of performance. However, if you did believe there is no difference in these algorithms, then your secondary criteria would probably be run time. KASBA is 30-50 times faster than DBA and MSM and 400 times faster than Shape-DBA. Run times are obviously implementation and hardware dependent but $k$-Shape and KASBA  are orders of magnitude faster than the others. There is little to split the two in terms of performance on data used to train the algorithm. However, on unseen data $k$-Shape does very poorly, whereas KASBA is the top performer. We think this result, and the fact KASBA is twice as fast as $k$-Shape on the combined datasets reported in Figure~\ref{fig:combined}, makes KASBA preferable to $k$-Shape as a baseline TSCL algorithm.

\subsection{KASBA against other partitional clusterers}
\label{sec:msm-results}
KASBA is an extension of the MSM with Barycentre Averaging (MBA) algorithm~\cite{holder23mba}. As discussed in Section~\ref{sec:tscl-algorithms}, the original MBA clusterer followed Lloyd’s algorithm, employing the MSM distance metric and the MSM elastic barycentre average. Additionally, it utilised Forgy initialisation with $10$ restarts.

Restarting 10 times obviously makes the algorithm 10 times slower. For our experiments with an MBA clusterer, we replace the initialisation algorithm with the elastic MSM $k$-means++ algorithm. This modification ensures that any improvements in clustering performance or run time arise independently of the initialisation method.

The primary aim of KASBA is to enhance the run time efficiency of MBA without compromising its clustering performance. An effective alternative to $k$-means that circumvents the challenges of averaging is the medoid-based PAM algorithm (see Section~\ref{sec:tscl-algorithms}). When combined with MSM, PAM has been shown to be a highly effective clustering approach~\cite{holder23kmedoids}.

\begin{figure}[htb]
    \centering
            \begin{tabular}{c c}
    \includegraphics[width=0.5\linewidth]{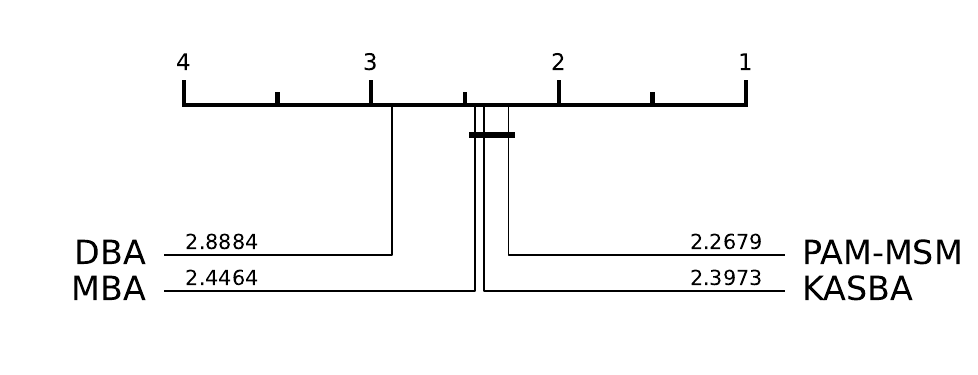} &
    \includegraphics[width=0.5\linewidth]
    {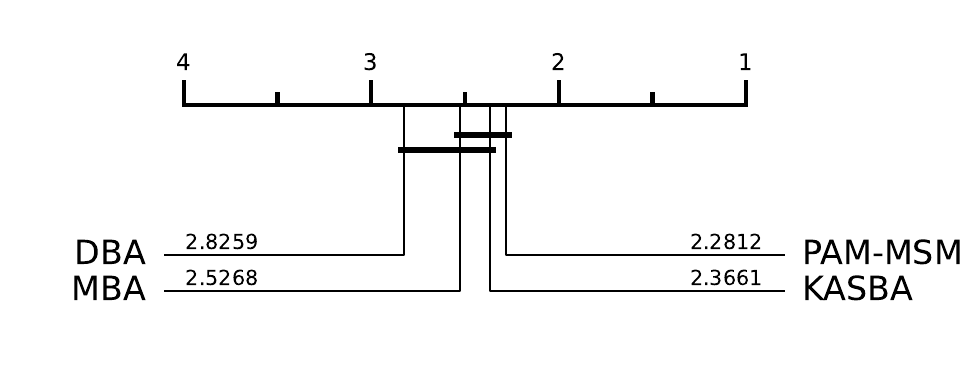}\\
    Accuracy & Adjusted Rand Index \\
    \includegraphics[width=0.5\linewidth]
    {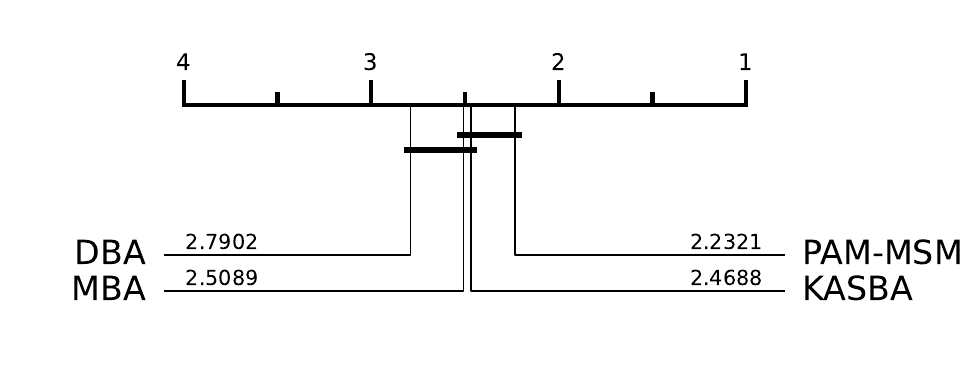} & 
    \includegraphics[width=0.5\linewidth]
    {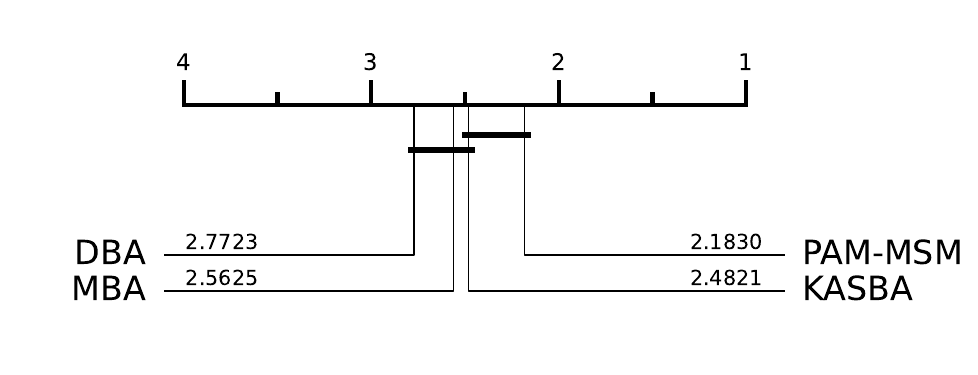} \\
    Average Mutual Information & Normalised Mutual Information \\
    \end{tabular}
    \caption{KASBA against $k$-means with DTW barycentre averaging (DBA), MSM barycentre averaging (MBA) and Partitioning Around the Medoid with MSM (pam-msm), ranks averaged over the 112 UCR archive data completed by all algorithms using the default test-train split.}
    \label{fig:msm}
\end{figure}
Figure~\ref{fig:msm} shows the performance of the three MSM variants, with DBA included for reference. KASBA is not significantly worse than the other two MSM based clusterers, but it is significantly faster. Table~\ref{tab:msm_times} summarises the run times. KASBA is 30 to 120 times faster. PAM requires the calculation of the pairwise distance matrix prior to clustering, which can add a huge time and memory cost for large problems; PAM-MSM takes seven days to cluster the largest problem in the archive, Crop. If we combine the train and test set, CROP has 240,000 unique time series. KASBA completes the clustering in minutes. PAM-MSM did not finish within the seven-day run time limit. PAM would require 288,000,000 unique elastic distance function calls and need over 2GB of memory to store the matrix.

\begin{table}[!ht]
    \vspace{0.25cm}
    \centering
    \begin{tabular}{|l|l|l|l|l|}
        \hline
Estimator  & Median (mins) & Mean (mins) & Max (hrs) & Total (hrs) \\ \hline
KASBA      & 0.0           & 2.1         & 1.2       & 3.9         \\
MBA        & 0.0           & 32.2        & 13.7      & 60.2        \\
DBA        & 1.2           & 55.0        & 22.9      & 102.7       \\
PAM-MSM    & 0.0           & 228.7       & 234.3     & 426.9       \\ \hline
    \end{tabular}
    \vspace{0.25cm}
    \caption{Summary of training times for four MSM based clustering algorithms on 112 UCR datasets completed. Timings are for a sequential training runs for all 112 datasets.}
    \label{tab:msm_times}
\end{table}

\subsection{KASBA against other TSCL algorithms}
\label{sec:others}
For context we compare KASBA to three clustering algorithms that are not distance based. We have selected these particular algorithms because they were published in good venues and we have implementations of them readily available. 

\begin{figure}[htb]
    \centering
            \begin{tabular}{c c}
    \includegraphics[width=0.5\linewidth]{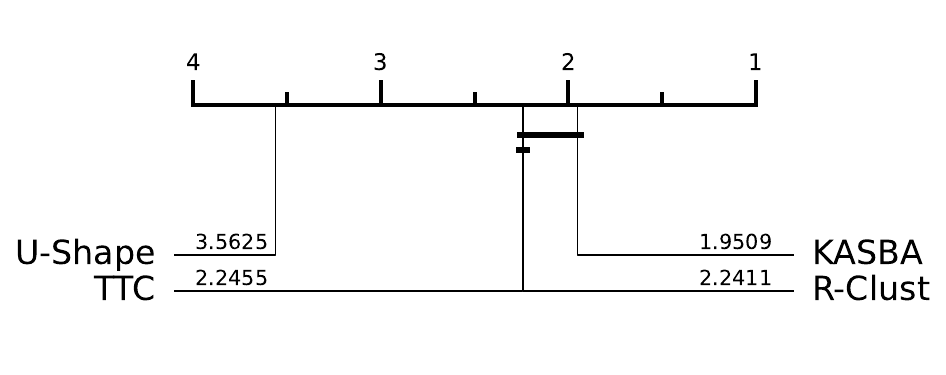} &
    \includegraphics[width=0.5\linewidth]
    {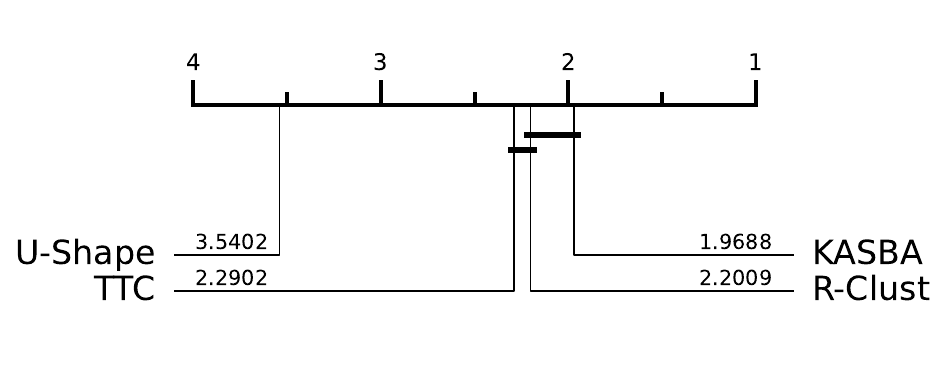}\\
    Accuracy & Adjusted Rand Index \\
    \includegraphics[width=0.5\linewidth]
    {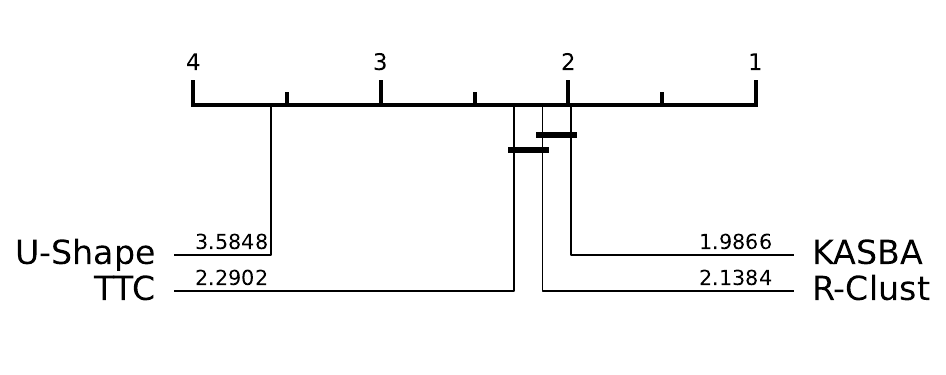} & 
    \includegraphics[width=0.5\linewidth]
    {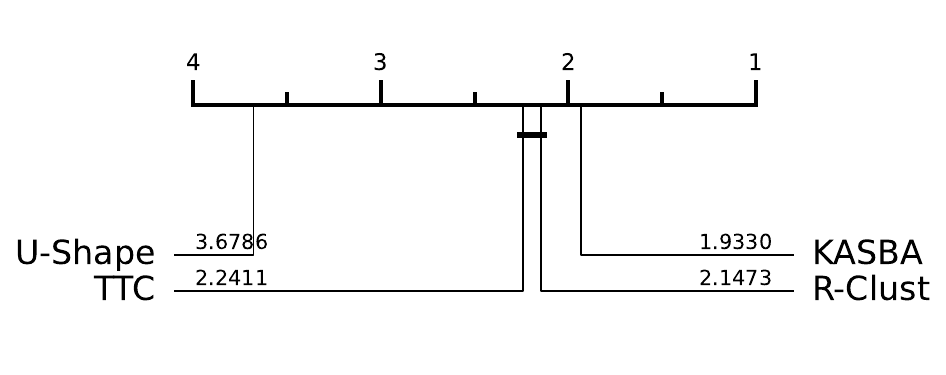} \\
    Average Mutual Information & Normalised Mutual Information \\
    \end{tabular}
    \caption{KASBA against alternative TSCL algorithms U-Shapelets, R-Clustering and TTC, with ranks averaged over the 112 UCR archive data completed by all algorithms using the default test-train split.}
    \label{fig:others}
\end{figure}
Figure~\ref{fig:others} shows that KASBA is significantly better or not significantly worse using all four measures, further supporting its case as use as a strong benchmark.

Results for $300$ different deep learning clustering algorithms on the same UCR datasets we use were presented in~\cite{lafabregue22deep}. We have not recreated these results due to computational constraints. However, the associated repository\footnote{https://github.com/blafabregue/TimeSeriesDeepClustering}  provides NMI results for over $300$ different clustering algorithms on the same default train/test splits of the UCR datasets. These are not directly comparable, since they are averaged over five runs and there may be other experimental differences. However, they can give some indication of relative performance. The best deep learning approach was a convolutional neural network with joint pretext loss and without clustering loss (key in their results is res\_cnn\_joint\_None). It achieved an average NMI of $0.3292$ over the 112 problems we use. KASBA average NMI is $0.3135$. The deep learning NMI are much higher on a two datasets, skewing the average. Overall, KASBA was better on 57 problems, the best deep learner best on 55. Given selecting the best of 300 is likely to bias the results, we believe this demonstrates that KASBA is competitive with deep learning TSCL algorithms while being significantly faster.

\section{Analysis of KASBA}
\label{sec:analysis}
To investigate the applicability of KASBA and analyse the importance of its design components, we conduct additional comparisons under controlled conditions. This section explores the impact of various factors, including initialisation techniques, distance measures, barycentre initialisation methods, average subset sizes for barycentre computation, and the utilisation of the metric property in conjunction with KASBA. Our goal is to highlight the importance of each component.

\subsection{Exploratory Analysis}
\begin{figure}[!tb]
    \centering
    \includegraphics[width=0.8\linewidth]  {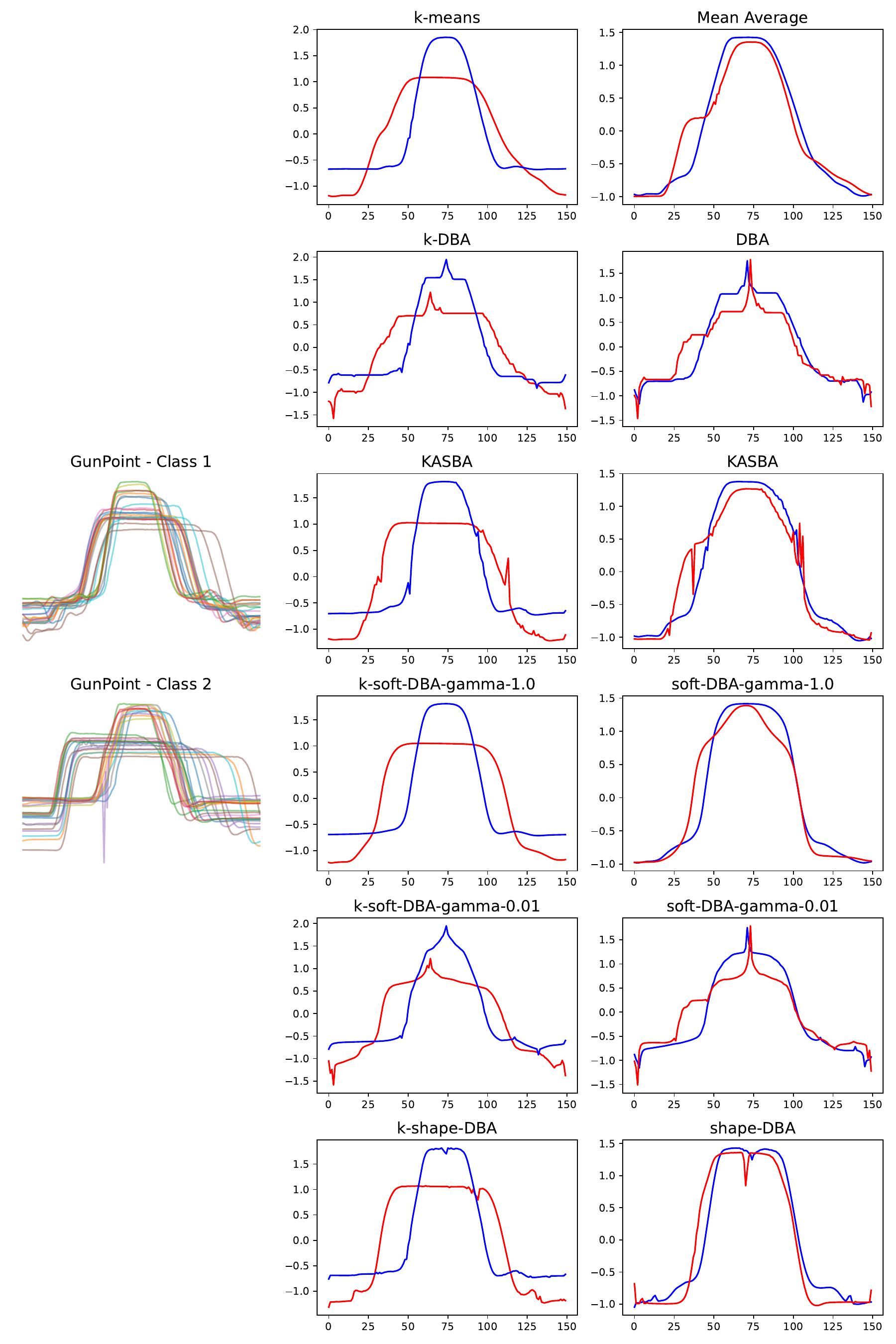} \caption{Different prototypes for GunPoint data. The left hand plots show the data, the middle column shows the prototypes formed using the class labels when using different averaging techniques. The column on the right shows the actual cluster centroids found by the respective algorithms.}
    \label{fig:gunpoint}
\end{figure}

Clustering is most commonly used as an exploratory tool, where a training and test split typically does not exist. When considering the combined train/test split we observe that similar algorithms such as DBA and shape-DBA are consistently in the top clique with KASBA. However, soft-DBA underperformed compared to our expectations. In previous studies~\cite{holder24tsclwithkmeans}, soft-DBA, when paired with Forgy initialisation and $10$ restarts, significantly outperformed the current state-of-the-art algorithms (e.g., DBA and shape-DBA). However, when using elastic $k$-means++, soft-DBA appears to perform significantly worse.

Initially, we hypothesised that the $\gamma$ parameter might explain this behaviour. The original soft-DBA paper~\cite{cuturi2017softdtw} states: {\em ``low smoothing parameter $\gamma$ yields barycenters that are spurious. On the other hand, a descent on the soft-DTW loss with sufficiently high $\gamma$ converges to a reasonable solution.''} We used $\gamma = 1.0$ because it is the value most likely to ensure convergence (other values were shown to fail to converge on multiple datasets~\cite{cuturi2017softdtw}). To verify this hypothesis, we conducted additional experiments with a different $\gamma$ value: $\gamma = 0.001$. However, we found no significant differences in overall performance across the tested $\gamma$ values.

To better understand and explain results, such as the poor performance of soft-DBA, and to highlight the exploratory power of KASBA, we compared various algorithms on the classic GunPoint problem~\cite{ye11shapelets}. This dataset records the movement along the x-axis of two actors performing two actions: drawing a gun from a holster at their belt and mimicking the same action without a gun, instead pointing their finger.

We applied various averaging methods to instances from each class using techniques from standard $k$-means (mean average), DBA, KASBA, Soft-DBA, and Shape-DBA. The central column of Figure~\ref{fig:gunpoint} presents the prototype for each class overlaid: Class 1 (gun) is shown in blue, and Class 2 (point) in red. The mean average produces similar shapes for both classes, with a notable difference between points 20–50, where the gun is being drawn from the holster in Class 1. The DBA prototypes exhibit peaks in the middle, albeit at different positions for each class, with Class 1 displaying a higher plateau. However, these prototypes are challenging to interpret in terms of motion. Soft-DBA appears similar to the mean average when $\gamma = 1.0$, but resembles DBA when $\gamma = 0.01$. Shape-DBA, on the other hand, inverts the peak visible in both DBA and Soft-DBA, resulting in distinct centroids. 

KASBA highlights distinctive variations around time points 40–60 and 100–120, which correspond to the moments when the hand is reaching for the gun and drawing it from the holster. These intervals are known to be the most discriminative for this dataset~\cite{ye11shapelets}. This example demonstrates how the centroids produced by KASBA can guide exploratory analysis toward key areas of interest in the time series.
We also extracted the actual centroids from a clustering run for these algorithms, as shown on the right in Figure~\ref{fig:gunpoint}. While all centroids resemble the class prototypes depicted in the middle of Figure~\ref{fig:gunpoint}, they appear more attenuated. Notably, KASBA preserves the variation around the gun-drawing motion, maintaining its discriminatory power.

\subsection{Ablative Study}

We introduce several novel features into the $k$-means clustering process to improve performance, hasten convergence and/or reduce run time. In this section we assess the impact of each component on KASBA performance in terms of quality and speed of clustering.

\subsubsection{Using the elastic distance in kmeans++ initialisation}
\begin{figure}[!htb]
    \centering
            \begin{tabular}{c c}
    \includegraphics[width=0.5\linewidth]{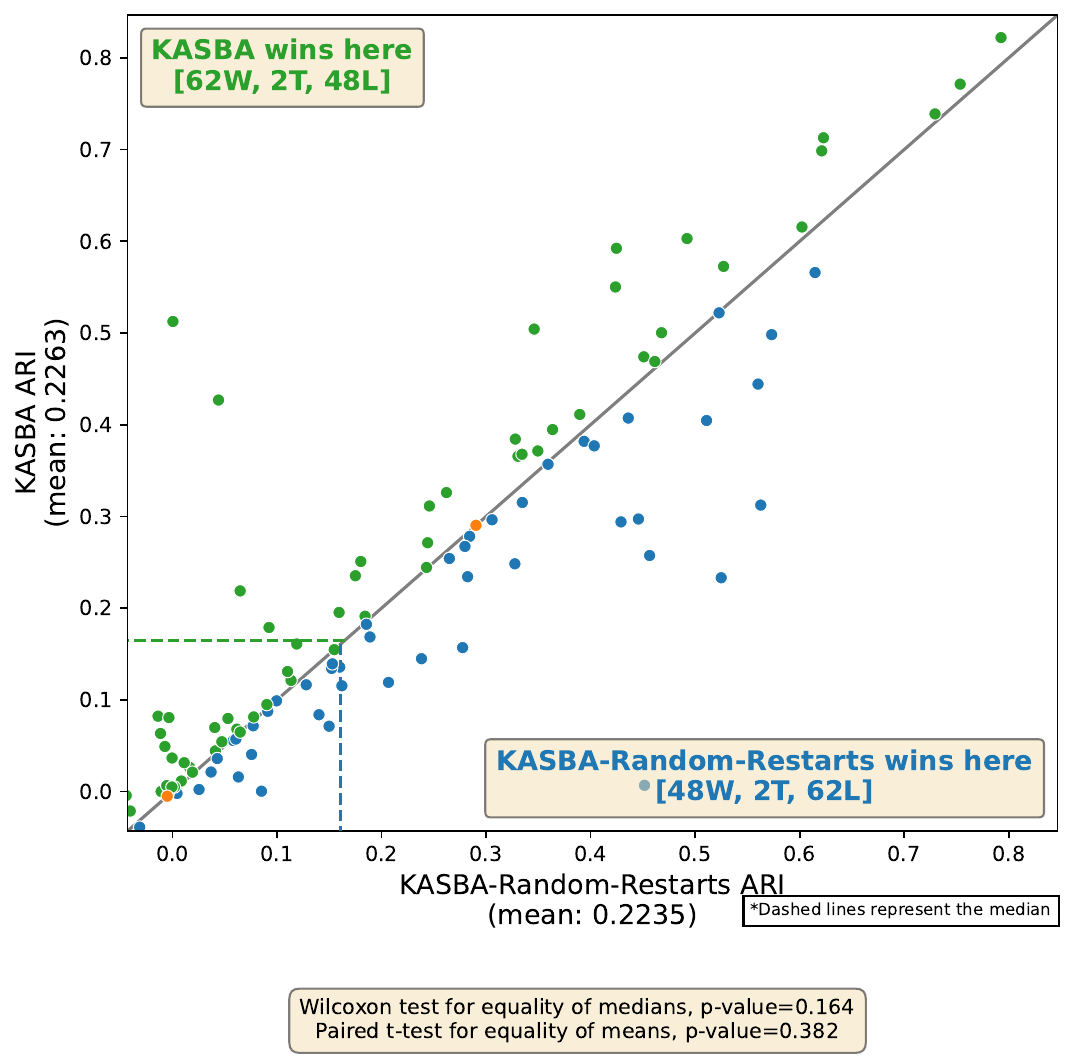} &
    \includegraphics[width=0.5\linewidth]
    {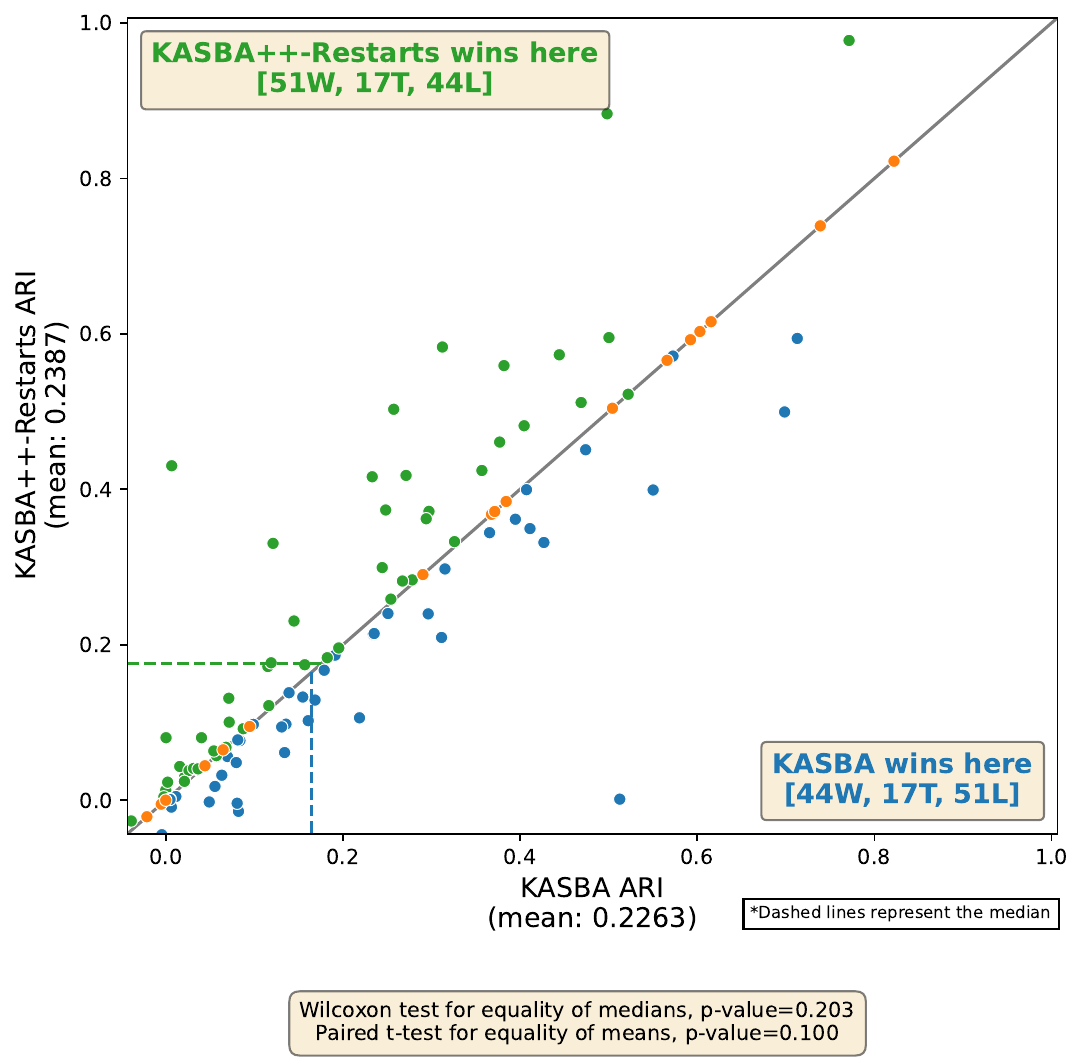}\\
    (a) & (b)
    \end{tabular}
    \caption{Scatter plot of ARI for single run KASBA against 10 random restarts and 10 kmeans++ restarts.}
    \label{fig:init}
\end{figure}

KASBA introduces elastic kmeans++ initialisation, described in Algorithm~\ref{algo:elastic-kmeans-plus-plus}. We have compared this to the standard approach of using Forgy with 10 restarts. The scatter plots for the ARI are shown Figure~\ref{fig:init}(a). There is no significant difference in performance, but random restarts is obviously an order of magnitude slower. We also tried restarting kmeans++ 10 times. Figure~\ref{fig:init}(b) shows there is a small benefit from restarting kmeans++, but it is not significant and we feel not worth the extra computation. It is easy to configure KASBA to do so if desired.

\subsubsection{Elastic barycentre averaging}
\begin{figure}[!htb]
    \centering
            \begin{tabular}{c c}
    \includegraphics[width=0.5\linewidth]
    {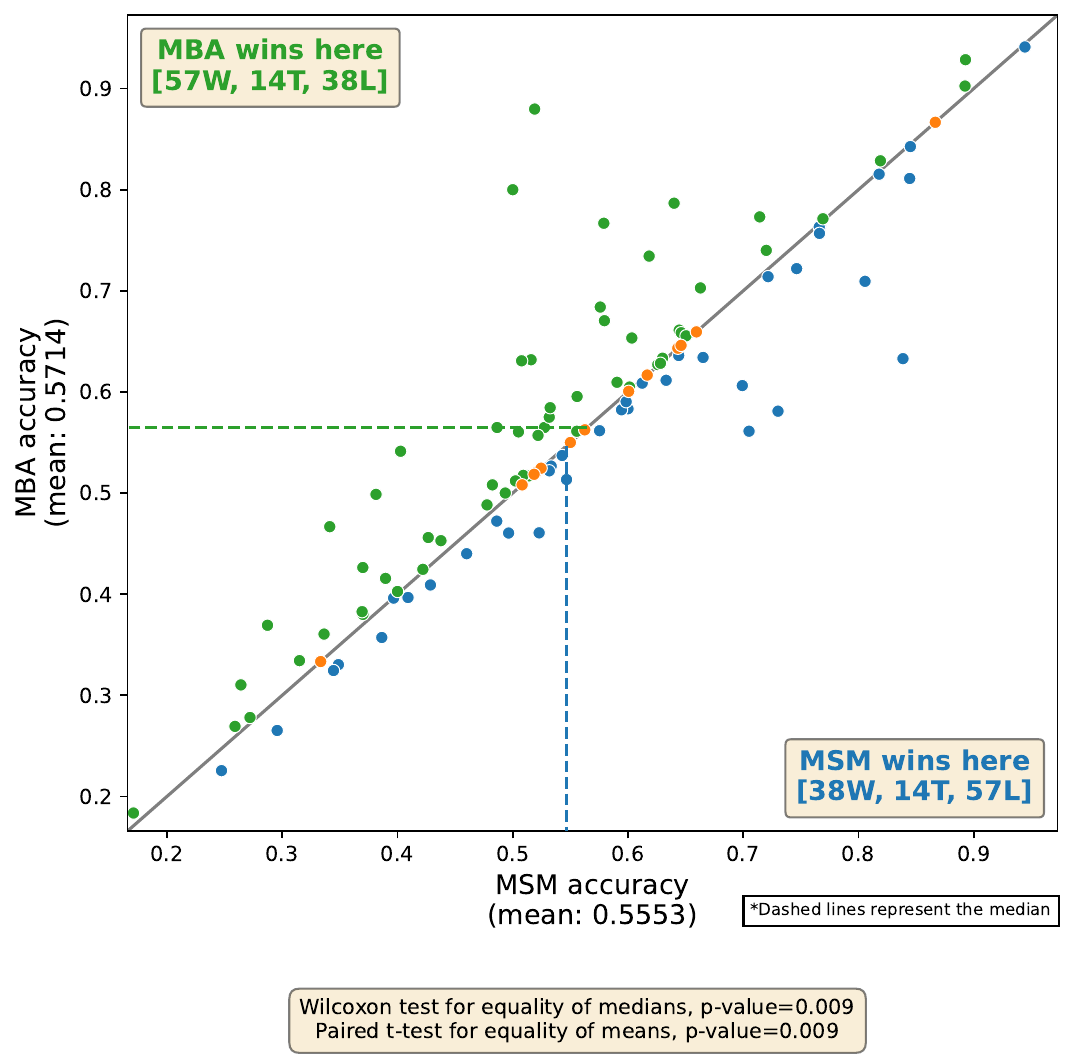} &
    \includegraphics[width=0.5\linewidth]
    {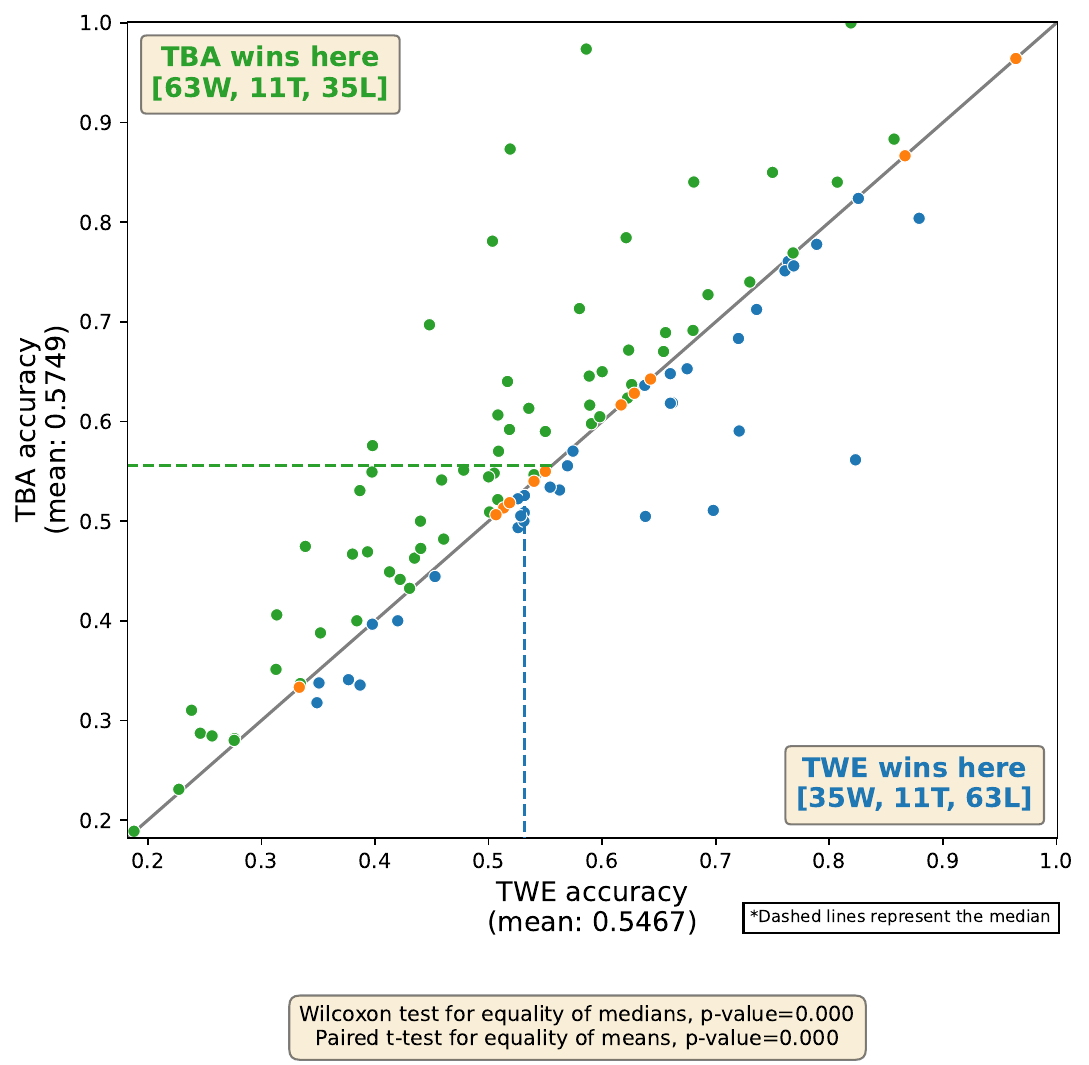}
    \end{tabular}
    \caption{Scatter plots of MSM/TWE with barycentre averaging (MBA/TBA) against $k$-means using MSM/TWE for asignment only (MSM/TWE) over 109 UCR archieve .}
    \label{fig:ba}
\end{figure}

\begin{figure}[htb]
    \centering
            \begin{tabular}{c c}
    \includegraphics[width=0.5\linewidth]{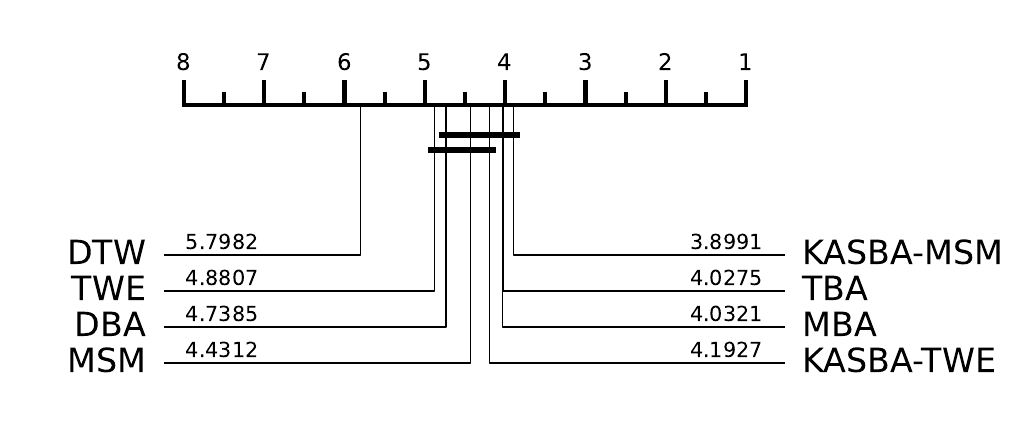} &
    \includegraphics[width=0.5\linewidth]
    {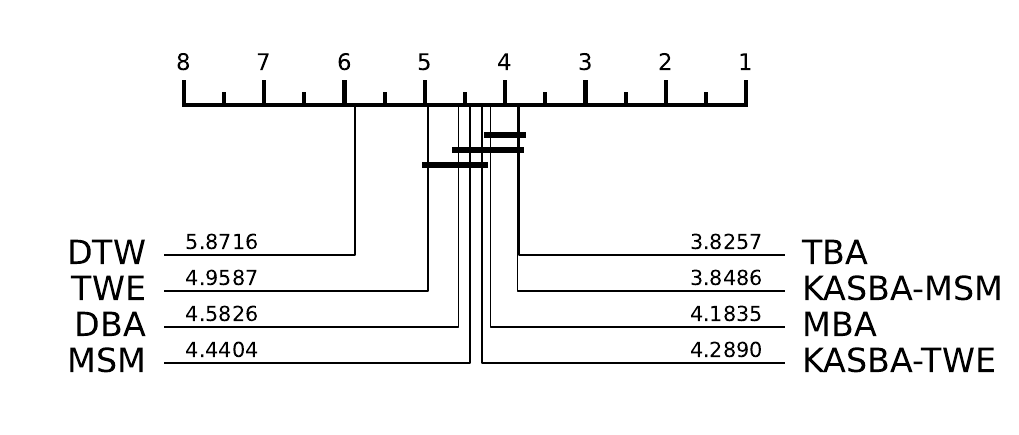}\\
    Accuracy & Adjusted Rand Index \\
    \includegraphics[width=0.5\linewidth]
    {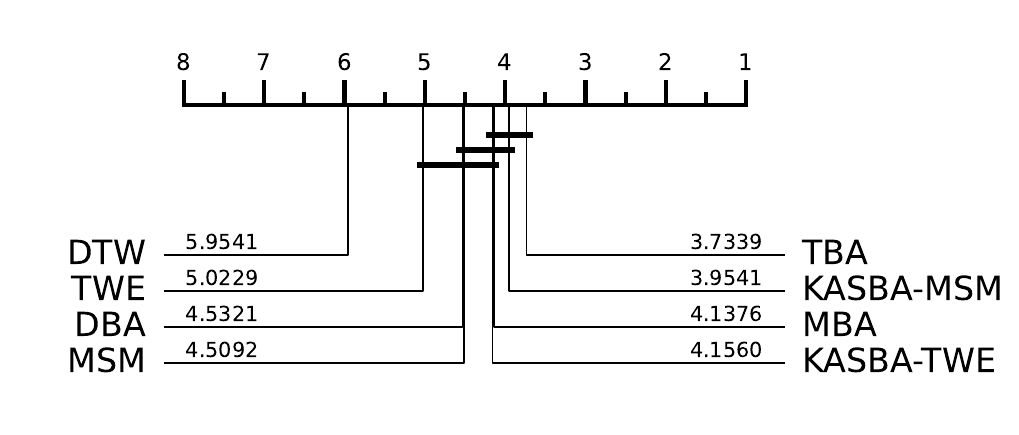} & 
    \includegraphics[width=0.5\linewidth]
    {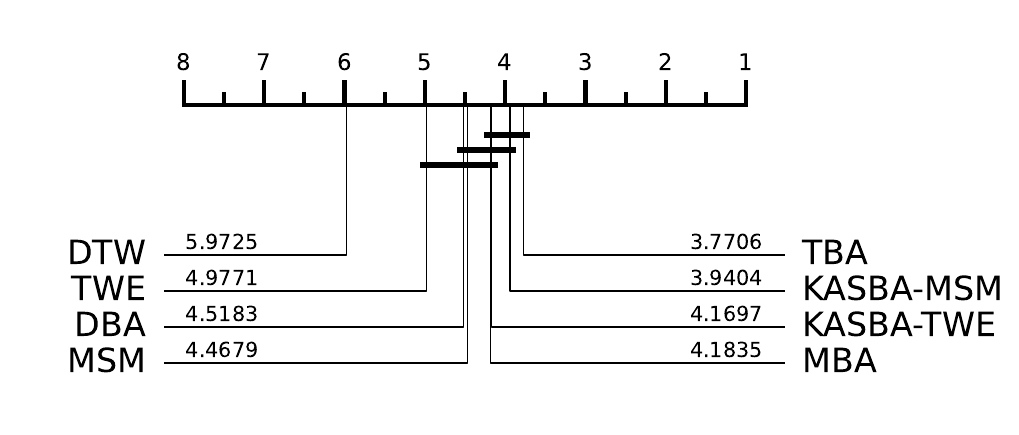} \\
    Average Mutual Information & Normalised Mutual Information \\
    \end{tabular}
    \caption{Elastic distance based $k$-means with ranks averaged over the 109 UCR archive data completed by all algorithms using the default test-train split.}
    \label{fig:dba2}
\end{figure}

The $k$-means algorithm uses distances in both its update and assignment stages. By contrast, DBA employs DTW in both stages, consistently outperforming $k$-means that uses DTW for assignment only~\cite{petitjean11dba}, a result corroborated in~\cite{holder24review}.

The idea of incorporating MSM for barycentre averaging (MBA), alongside its use for assignment, was introduced in~\cite{holder23mba}. While our results focus on the MSM distance, an alternative elastic distance function that is also a metric is the Time-Warp Edit (TWE) distance~\cite{marteau09twe}. Elastic barycentre averaging can also be applied with TWE, yielding a clusterer we denote TBA. 
 We refer to $k$-means algorithms that use an elastic distance solely during assignment as DTW, TWE, or MSM, depending on the specific distance used.
Figure~\ref{fig:ba} presents scatter plots comparing MBA against MSM and TBA against TWE, demonstrating that both barycentre-based approaches outperform $k$-means with elastic distance applied only during assignment.

MBA and TWE also significantly outperform DTW and DBA, as illustrated in Figure~\ref{fig:dba2}, which displays the critical difference diagram for all eight clusterers. TBA, MBA, KASBA-TWE and KASBA-MSM consistently rank in the top clique across all evaluation metrics. Specifically, for AMI, NMI, and ARI, they are significantly better than MSM, DBA, and TWE. While MBA and TBA consistently share the top clique with KASBA-MSM and KASBA-TWE, as detailed in Section~\ref{sec:results}, they are significantly slower than KASBA.

\subsubsection{Seeding of the stochastic subgradient descent with the previous centroid} 

\begin{figure}[htb]
    \centering
            \begin{tabular}{c c}
    \includegraphics[width=0.5\linewidth]{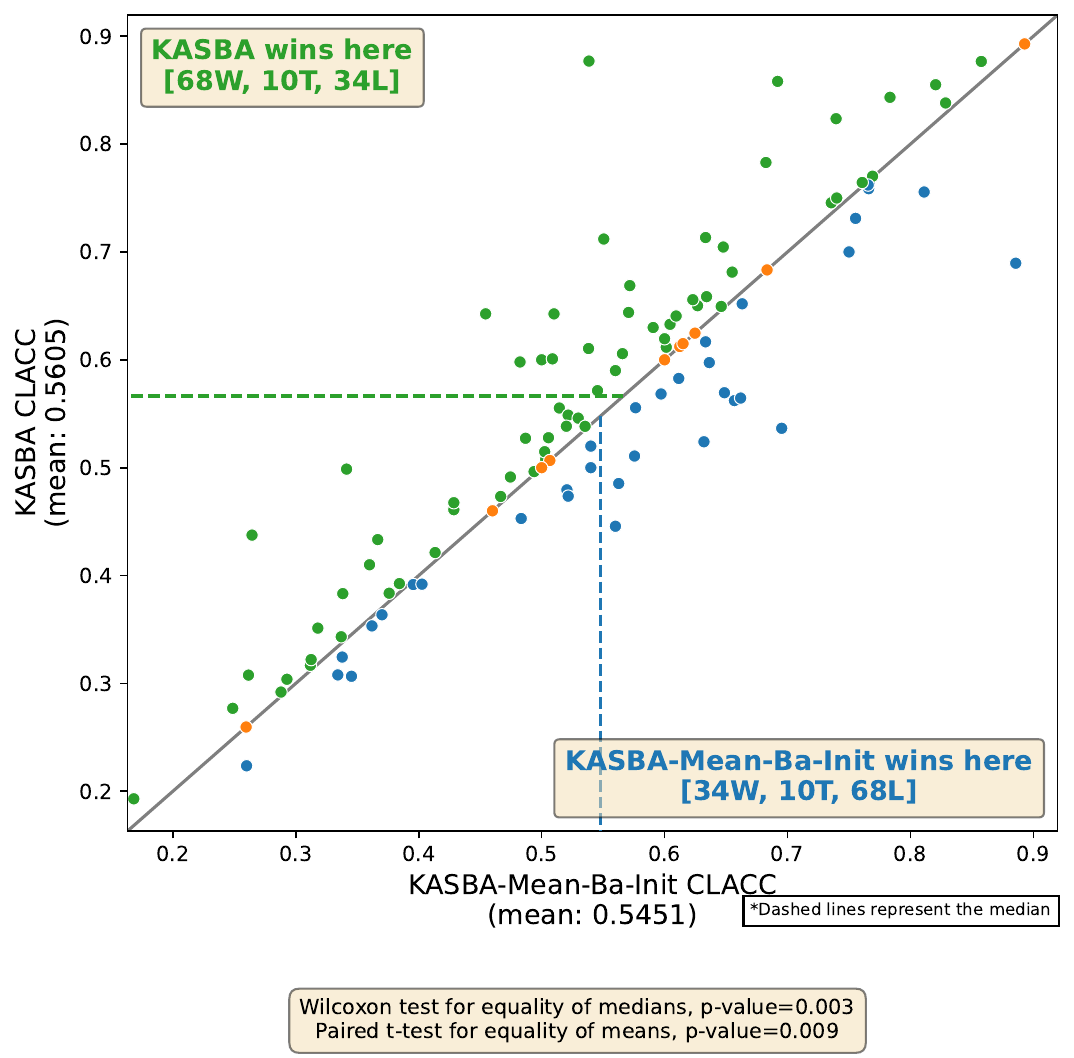} &
    \includegraphics[width=0.5\linewidth]
    {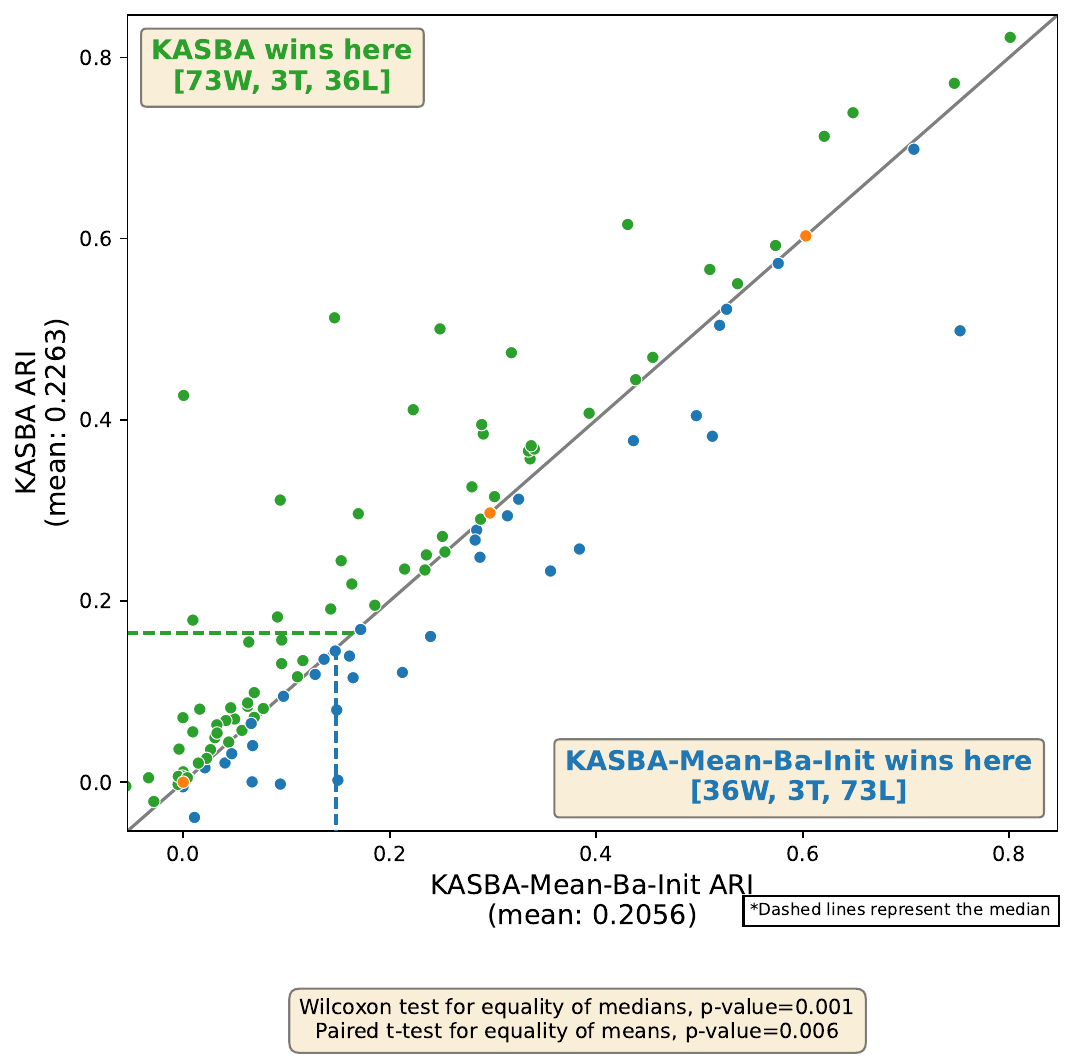}\\
    \end{tabular}
    \caption{Scatter plot for clustering accuracy (CLACC) and Adjusted Rand Index (ARI) of KASBA using centroid seeding to initialise gradient descent (KASBA) and KASBA using arithmetic mean (KASBA-MEAN).}
    \label{fig:seed}
\end{figure}

We use the centroid found at the last iteration (or at initialisation) as a starting point for the search for a new centroid rather than the arithmetic mean proposed in~\cite{petijean16faster} (see Algorithm~\ref{algo:elastic_ssg} line 8). 
This refinement is only apparent when you consider the clustering algorithm as a whole process rather than averaging in isolation: it improves both run time and performance. Seeding makes KASBA take on average 80\% of time taken when initialising with the average. This is achieved through making 60\% of the distance calls in update (4237 calls compared to 7210). It also results in better clusters.  Figure~\ref{fig:seed} shows the performance of KASBA with and without seeding.

Seeding the centroid allows us to reuse the sum of distances to the centroid (parameter $dist$ in Algorithms~\ref{algo:elastic_ssg}), which saves distance calculations and makes the algorithm more stable. To demonstrate this, we reran KASBA where instead of passing the distance sum, we always performed the first epoch of the gradient descent. We found this massively increased the number of epochs it took to reach convergence, rising from an average of 3.4 to 16.5, with a commensurate huge increase in distance calls and hence run time.

\subsubsection{Random subset gradient descent}

When finding a barycentre with batch stochastic subgradient descent (Algorithm~\ref{algo:elastic_ssg}) we take a random subset of series in the cluster to update the centroid on all epochs after the first. This is meant to reduce the number of distance calls on any epoch. However, it is also meant to decrease the likelihood of premature convergence by injecting some diversity into the process. We would expect subsampling to increase the number of epochs. This is desirable, to counteract the seeding risking premature convergence, but it does also increase the number of distance calls.

\begin{figure}[htb]
    \centering
            \begin{tabular}{c c}
    \includegraphics[width=0.5\linewidth]{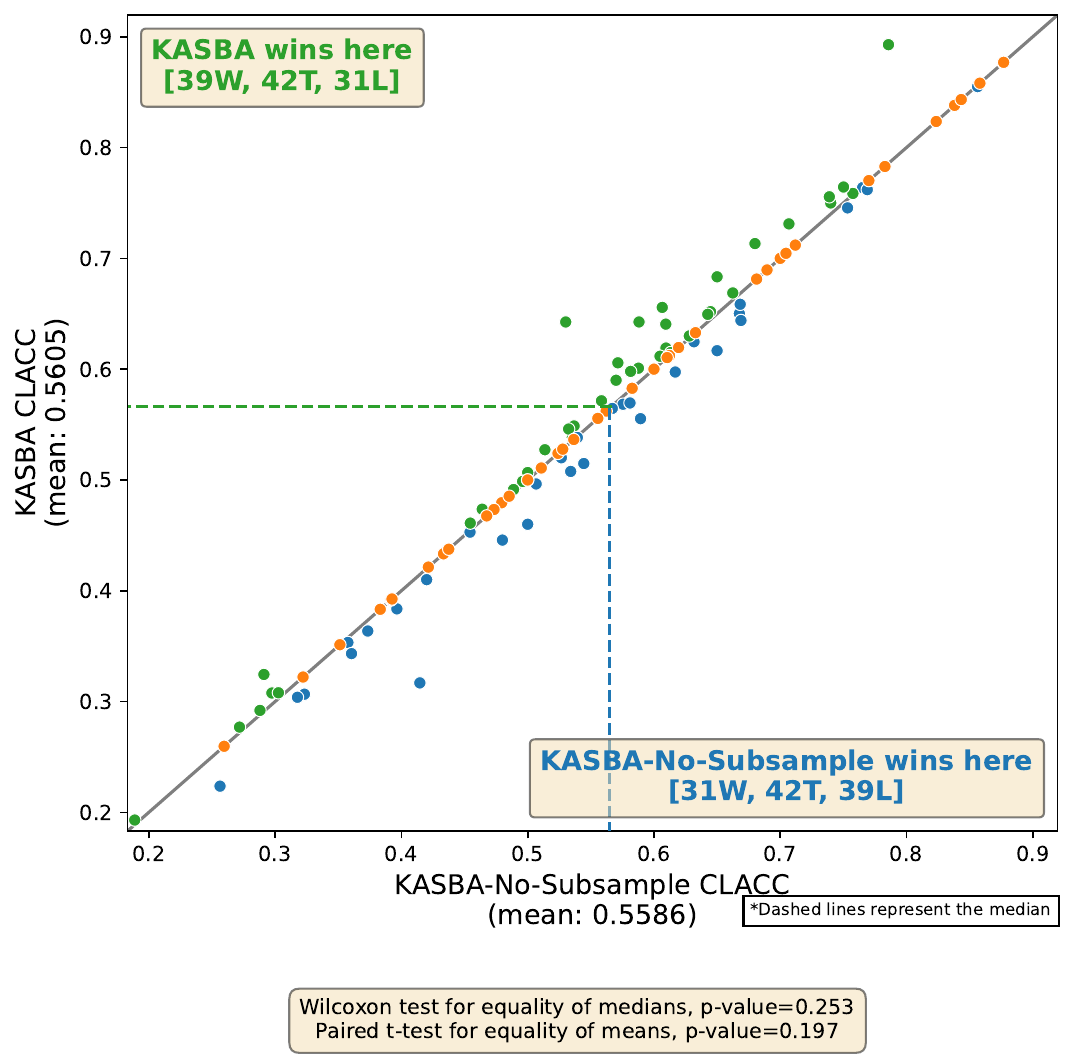} &
    \includegraphics[width=0.5\linewidth]
    {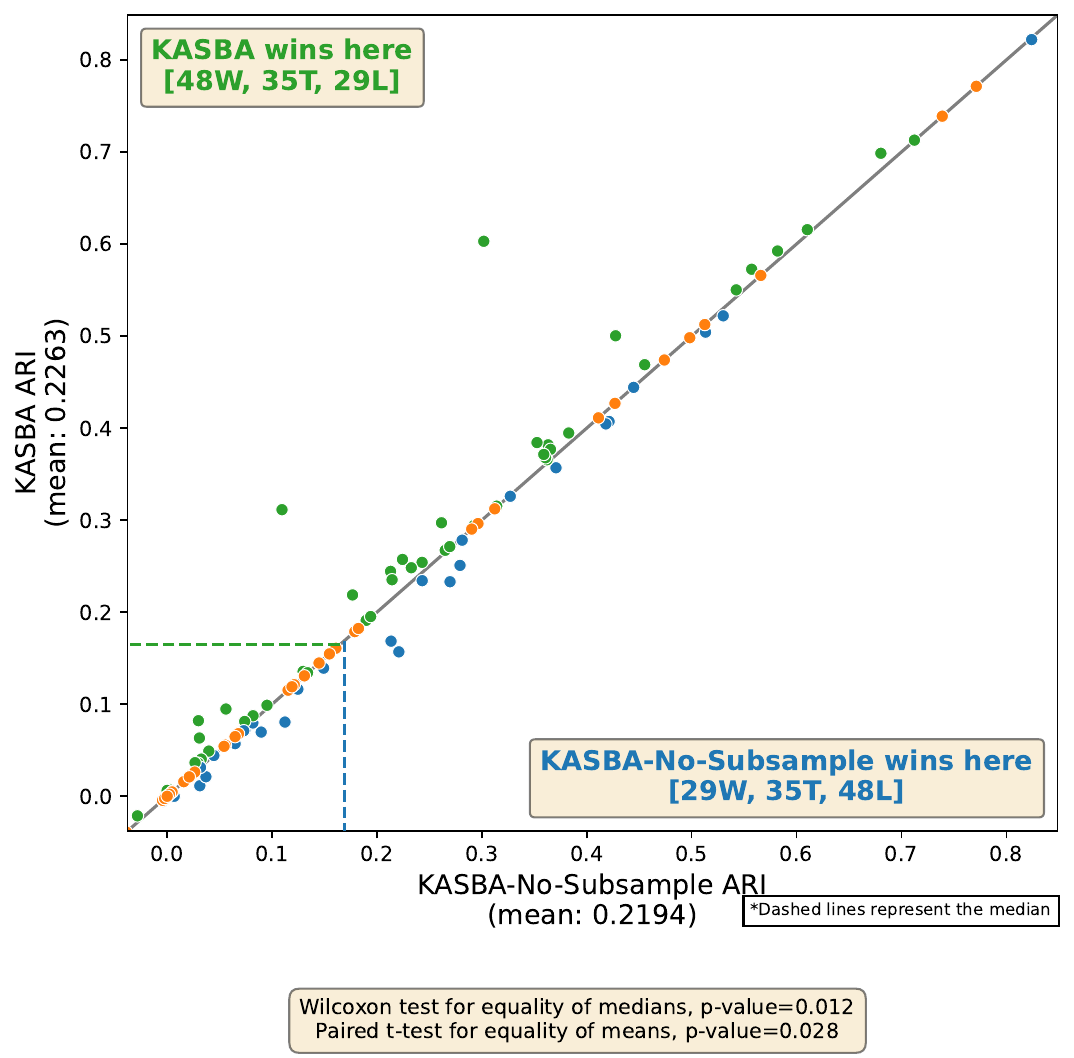}\\
    \end{tabular}
    \caption{Scatter plot of KASBA  using subsampling in the gradient descent against using the full cluster membership for each epoch (KASBA-Full).}
    \label{fig:fullsample}
\end{figure}

To assess the impact of these competing factors, we reran KASBA using the whole dataset on every barycentre iteration. Figure~\ref{fig:fullsample} shows the scatter plot of using the full sample on all iterations vs random subsamples. In terms of accuracy (left plot), there is no significant difference. However, sampling produces significantly better ARI (right plot).

Overall, KASBA with subsampling is only marginally faster than using the full dataset in every iteration, even though it makes half as many distance calls per epoch. This indicates that it requires more epochs to converge than the whole dataset every iteration. To confirm this, we reran the experiments and tracked both the number of distance calls and epochs. Subsampling resulted in a total of 85\% of the distance calls during updates but required 10\% more epochs to reach convergence.

Although we initially hypothesised that subsampling would significantly reduce run time, the actual reduction was modest. However, it did lead to improved performance. We hypothesise that this improvement arises because using a random subset helps prevent premature convergence.

\subsubsection{Triangle Inequality in assignment}

We exploit the triangle inequality to speed up KASBA in Algorithm~\ref{algo:fast_label}, lines 8-10. To assess the impact of this, and to verify it has no impact on the actual final clustering, we rerun experiments with the test of whether to skip turned off. Rather than measure run time, we count the number of distance function calls. This has the added benefit of profiling the algorithm, since the vast majority of the computation is in the distance calls. Overall, most of the distance calculations happen in the assignment stage. KASBA makes on average 38,130 distance calls per problem, and these are split between initialisation: 5780 calls (15\%); update: 4237 calls (11\%);  and assignment: 28113 calls (74\%). If we do not apply the triangle inequality in assignment, we perform on average 66,005 distance calls: on average, the triangle inequality results in about 40\% of the distance calculations. Overall, KASBA is approximately twice as fast when using the triangle inequality, with identical results. 

\section{Conclusion}
\label{sec:conclusions}
We have presented the $k$-means (K) accelerated (A) Stochastic subgradient (S) Barycentre (B) Average (A) (KASBA) clustering algorithm. Our extensive experimentation demonstrates that KASBA performs at least as well as the current state-of-the-art clusterers, if not better, whilst requiring orders of magnitude less computation: we can cluster all 112 UCR datasets with KESBA in under two hours on an Apple M4 Max CPU.  

TSCL research with $k$-means often focusses on a single element of the partitional clustering such as the update/averaging stage. KASBA is designed with bespoke versions of the initialisation, update and assignment phases and their linkage is one of the reasons KASBA is so fast. We employ the same distance measure, MSM, throughout the algorithm, including initialisation with elastic $k$-means++. We propose a novel random subset stochastic subgradient descent averaging algorithm and exploit the metric property of MSM to make assignment faster. We improve convergence by retaining information between iterations. 

Our experiments demonstrate that KASBA’s performance is comparable to both PAM-MSM and the MBA whilst being two orders of magnitude faster. We also recreated results for three non distance based clusterers and compared to published results to the best deep learning algorithm. None of these outperformed KASBA.
KASBA has a small memory footprint and is very fast. It has few parameters, is simple and works well without tuning. We believe KASBA is a valuable addition to the TSCL cannon, a useful benchmark for future research and useful for practitioners.

\backmatter

\bmhead{Acknowledgements}

This work has been supported by the UK Research and Innovation Engineering and Physical Sciences Research Council (grant reference EP/W030756/2). Some of the experiments were carried out on the High Performance Computing Cluster supported by the Research and Specialist Computing Support service at the University of East Anglia. We would like to thank all those responsible for helping maintain the time series classification archives and those contributing to open source implementations of the algorithms.

\begin{appendices}

\section{Missing results}
\label{app:missing-datasets}
Results missing from the train/test results by clusterer are listed in Table~\ref{tab:test-train-split-missing-datasets}. Those missing from the combined results are listed in Table~\ref{tab:combined-test-train-split-missing-datasets}. A \checkmark means the dataset is present for the model and a x means the dataset is missing for the model. If a model is not included as a column it means all 112 datasets completed.
\begin{table}[htb]
\centering
\caption{Missing datasets for the train/test split experiments. A total of 14 datasets are excluded. Missing results for MSM were due to repeated empty cluster formation. Missing datasets for shape-DBA and soft-DBA was due to the run time exceeding our $7$ day run time limit.}
\label{tab:test-train-split-missing-datasets}
\begin{tabular}{|l|p{2cm}|p{2cm}|p{2cm}|p{2cm}|}
\hline
\bf{Dataset} & \bf{MSM} & \bf{shape-DBA} & \bf{soft-DBA} \\
\hline
EOGHorizontalSignal & \checkmark & \checkmark & x \\
\hline
EOGVerticalSignal & \checkmark & \checkmark & x \\
\hline
EthanolLevel & \checkmark & x & \checkmark \\
\hline
FordA & \checkmark & \checkmark & x \\
\hline
FordB & \checkmark & \checkmark & x \\
\hline
NonInvasiveFetalECGThorax1 & \checkmark & x & x \\
\hline
NonInvasiveFetalECGThorax2 & \checkmark & \checkmark & x \\
\hline
Phoneme & \checkmark & \checkmark & x \\
\hline
PigAirwayPressure & \checkmark & \checkmark & x \\
\hline
PigArtPressure & x & \checkmark & \checkmark \\
\hline
SemgHandMovementCh2 & \checkmark & \checkmark & x \\
\hline
SemgHandSubjectCh2 & \checkmark & \checkmark & x \\
\hline
ShapesAll & \checkmark & \checkmark & x \\
\hline
UWaveGestureLibraryAll & \checkmark & \checkmark & x \\
\hline
\bf{Total Missing} & \bf{1} & \bf{2} & \bf{12} \\
\hline
\end{tabular}
\end{table}

\begin{table}[t!bh]
\centering
\caption{Missing datasets for the combined train/test split experiments. A total of 19 datasets are excluded. Missing results for MSM and DBA were due to repeated empty cluster formation. Missing datasets for shape-DBA and $k$-SC was due to the run time exceeding our $7$ day run time limit.}
\label{tab:combined-test-train-split-missing-datasets}
\begin{tabular}{|l|p{2cm}|p{2cm}|p{2cm}|p{2cm}|p{2cm}|}
\hline
\bf{Dataset} & \bf{DBA} & \bf{MSM} & \bf{k-SC} & \bf{shape-DBA} \\
\hline
CinCECGTorso & \checkmark & \checkmark & \checkmark & x \\
\hline
EOGHorizontalSignal & \checkmark & \checkmark & \checkmark & x \\
\hline
EthanolLevel & \checkmark & \checkmark & \checkmark & x \\
\hline
FordA & \checkmark & \checkmark & \checkmark & x \\
\hline
FordB & \checkmark & \checkmark & \checkmark & x \\
\hline
HandOutlines & x & \checkmark & \checkmark & \checkmark \\
\hline
InlineSkate & \checkmark & \checkmark & \checkmark & x \\
\hline
MixedShapesRegularTrain & \checkmark & \checkmark & \checkmark & x \\
\hline
MixedShapesSmallTrain & \checkmark & \checkmark & \checkmark & x \\
\hline
NonInvasiveFetalECGThorax1 & \checkmark & \checkmark & \checkmark & x \\
\hline
Phoneme & \checkmark & \checkmark & \checkmark & x \\
\hline
PigArtPressure & \checkmark & x & \checkmark & \checkmark \\
\hline
PigCVP & \checkmark & x & \checkmark & \checkmark \\
\hline
SemgHandGenderCh2 & \checkmark & \checkmark & \checkmark & x \\
\hline
SemgHandMovementCh2 & \checkmark & \checkmark & \checkmark & x \\
\hline
SemgHandSubjectCh2 & \checkmark & \checkmark & \checkmark & x \\
\hline
StarLightCurves & \checkmark & \checkmark & \checkmark & x \\
\hline
UWaveGestureLibraryAll & \checkmark & \checkmark & \checkmark & x \\
\hline
UWaveGestureLibraryZ & \checkmark & \checkmark & x & \checkmark \\
\hline
\bf{Total Missing} & \bf{1} & \bf{2} & \bf{1} & \bf{15} \\
\hline
\end{tabular}
\end{table}
\end{appendices}

\newpage

\bibliography{sn-bibliography, TSCMaster}

\end{document}